\documentclass[10pt,conference]{IEEEtran}
\IEEEoverridecommandlockouts
\usepackage{cite}
\usepackage{amsmath,amssymb,amsfonts}
\usepackage{textcomp}
\usepackage{xcolor}
\usepackage{colortbl}
\usepackage{amsmath}
\usepackage[linesnumbered,ruled,vlined]{algorithm2e}
\usepackage{hyperref}
\hypersetup{hidelinks}
\usepackage{multirow}
\usepackage{bm}
\usepackage[normalem]{ulem}

\usepackage{tikz}
\usetikzlibrary{shapes}
\usetikzlibrary{positioning}
\usetikzlibrary{plotmarks}
\usetikzlibrary{patterns}
\usetikzlibrary{fit,calc}
\usepackage{pgfplots}
\usetikzlibrary{arrows.meta}
\usepackage[customcolors,shade]{hf-tikz}
\usepackage{microtype}
\usepackage{graphicx}
\usepackage{enumitem}
\usepackage{booktabs}
\usepackage{adjustbox}
\usepackage{xspace}
\usepackage{multirow}
\usepackage{tabularx}
\usepackage{tikz}
\usepackage{CJKutf8}
\usepackage{subcaption}
\usepackage{balance}
\usepackage{tikz}
\usetikzlibrary{positioning}
\usepackage{array} 

\definecolor{SystemColor}{RGB}{255, 230, 204}
\definecolor{ApplicationColor}{RGB}{204, 255, 204}
\definecolor{ModelsDataColor}{RGB}{204, 204, 255}

\usepackage{thmtools, thm-restate}
\ifCLASSOPTIONcompsoc
   
\else
 
\fi

\def\BibTeX{{\rm B\kern-.05em{\sc i\kern-.025em b}\kern-.08em
    T\kern-.1667em\lower.7ex\hbox{E}\kern-.125emX}}

\pgfplotsset{compat=1.18}
\begin{document}
\begin{CJK}{UTF8}{gbsn}
\title{
Towards General Industrial Intelligence: A Survey of Continual Large Models in Industrial IoT}

\author{
\IEEEauthorblockN{
Jiao Chen,
Jiayi He,
Fangfang Chen,
Zuohong Lv, 
Jianhua Tang,~\IEEEmembership{Senior~Member,~IEEE,}\\
Weihua Li,~\IEEEmembership{Senior~Member,~IEEE,} Zuozhu Liu,~\IEEEmembership{Member,~IEEE,} Howard H. Yang,~\IEEEmembership{Member,~IEEE,} \\and Guangjie Han,~\IEEEmembership{Fellow,~IEEE}
}
\thanks{This research work has been submitted to IEEE for peer review and potential publication. Should the copyright be transferred, this version may become inaccessible without prior notice.}
\thanks{
    Jiao Chen and Jiayi He contributed equally to this work. The corresponding author is Jianhua Tang.
}
\thanks{
    Jiao Chen, Jiayi He, Fangfang Chen, Zuohong Lv, and Jianhua Tang are with the Shien-Ming Wu School of Intelligent Engineering, South China University of Technology, Guangzhou 511442, China. Weihua Li is with the School of Mechanical and Automotive Engineering, South China University of Technology, Guangzhou 510641, China. Jianhua Tang and Weihua Li are also with Pazhou Lab, Guangzhou 510335, China. Howard H. Yang and Zuozhu Liu are with the ZJU-UIUC Institute, Zhejiang University, Haining 314400, China. Guangjie Han is with the Key Laboratory of Maritime Intelligent Network Information Technology, Ministry of Education, Hohai University, Changzhou 213200, China. 
    Email addresses: \{202110190459, 202164020171, wifannychen59, 202220159664\}@mail.scut.edu.cn, jtang4@e.ntu.edu.sg, whlee@scut.edu.cn, \{haoyang, zuozhuliu\}@intl.zju.edu.cn, hanguangjie@gmail.com.
}
}
\maketitle

\begin{abstract}
Industrial AI is transitioning from traditional deep learning models to large-scale transformer-based architectures, with the Industrial Internet of Things (IIoT) playing a pivotal role. IIoT evolves from a simple data pipeline to an intelligent infrastructure, enabling and enhancing these advanced AI systems.
This survey explores the integration of IIoT with large models (LMs) and their potential applications in industrial environments.
We focus on four primary types of industrial LMs: language-based, vision-based, time-series, and multimodal models. The lifecycle of LMs is segmented into four critical phases: data foundation, model training, model connectivity, and continuous evolution.
First, we analyze how IIoT provides abundant and diverse data resources, supporting the training and fine-tuning of LMs. Second, we discuss how IIoT offers an efficient training infrastructure in low-latency and bandwidth-optimized environments. Third, we highlight the deployment advantages of LMs within IIoT, emphasizing IIoT's role as a connectivity nexus fostering emergent intelligence through modular design, dynamic routing, and model merging to enhance system scalability and adaptability. Finally, we demonstrate how IIoT supports continual learning mechanisms, enabling LMs to adapt to dynamic industrial conditions and ensure long-term effectiveness.
This paper underscores IIoT's critical role in the evolution of industrial intelligence with large models, offering a theoretical framework and actionable insights for future research.
\end{abstract}

\section{Introduction}\label{sec:intro}
\begin{figure}[t]
    \centering
    \includegraphics[width=0.98\linewidth]{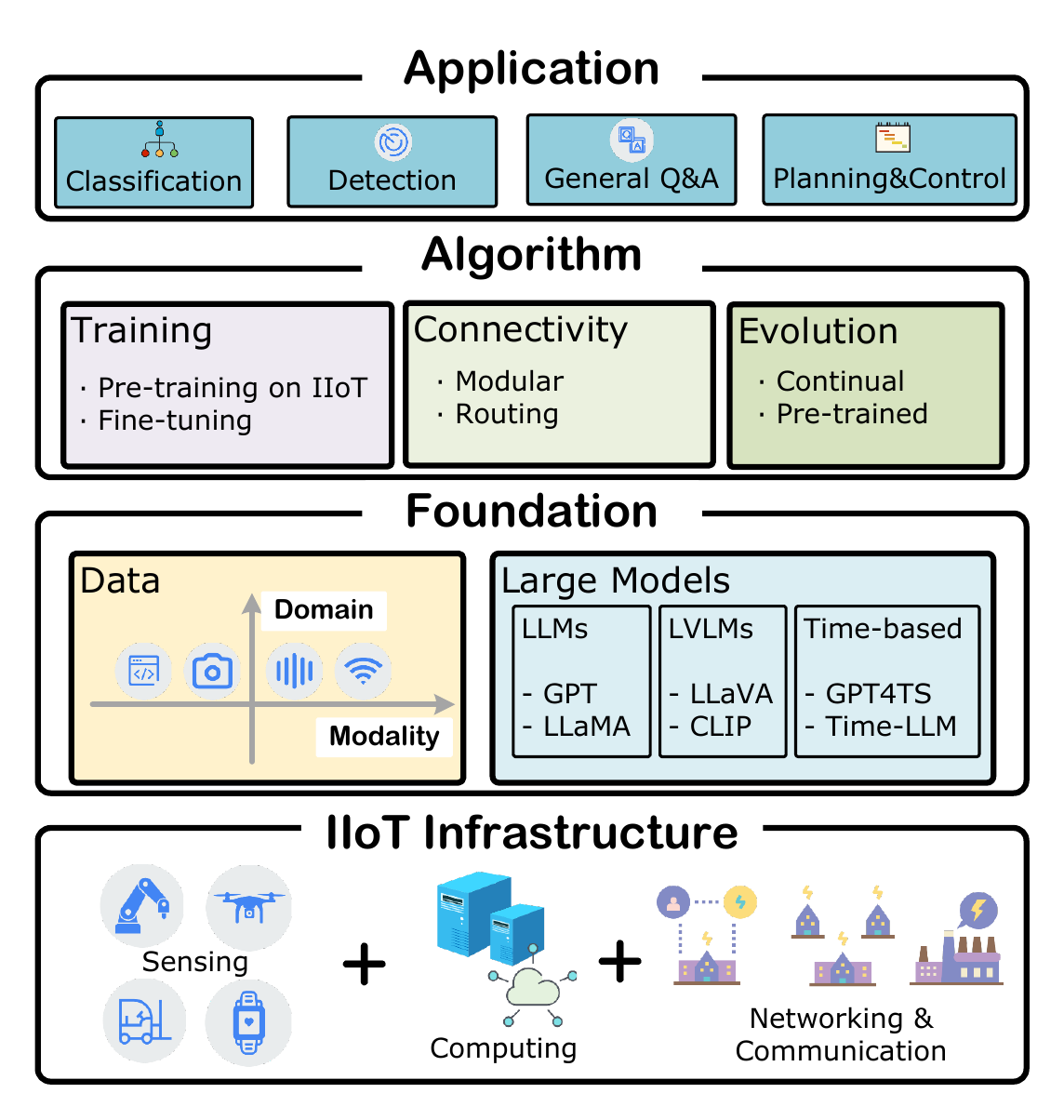}
    \caption{
    Layering the general industrial intelligence ecosystem.
    }
    \label{fig:gii}
\end{figure}

\begin{figure*}[t]
    \centering
    \includegraphics[width=0.8\linewidth]{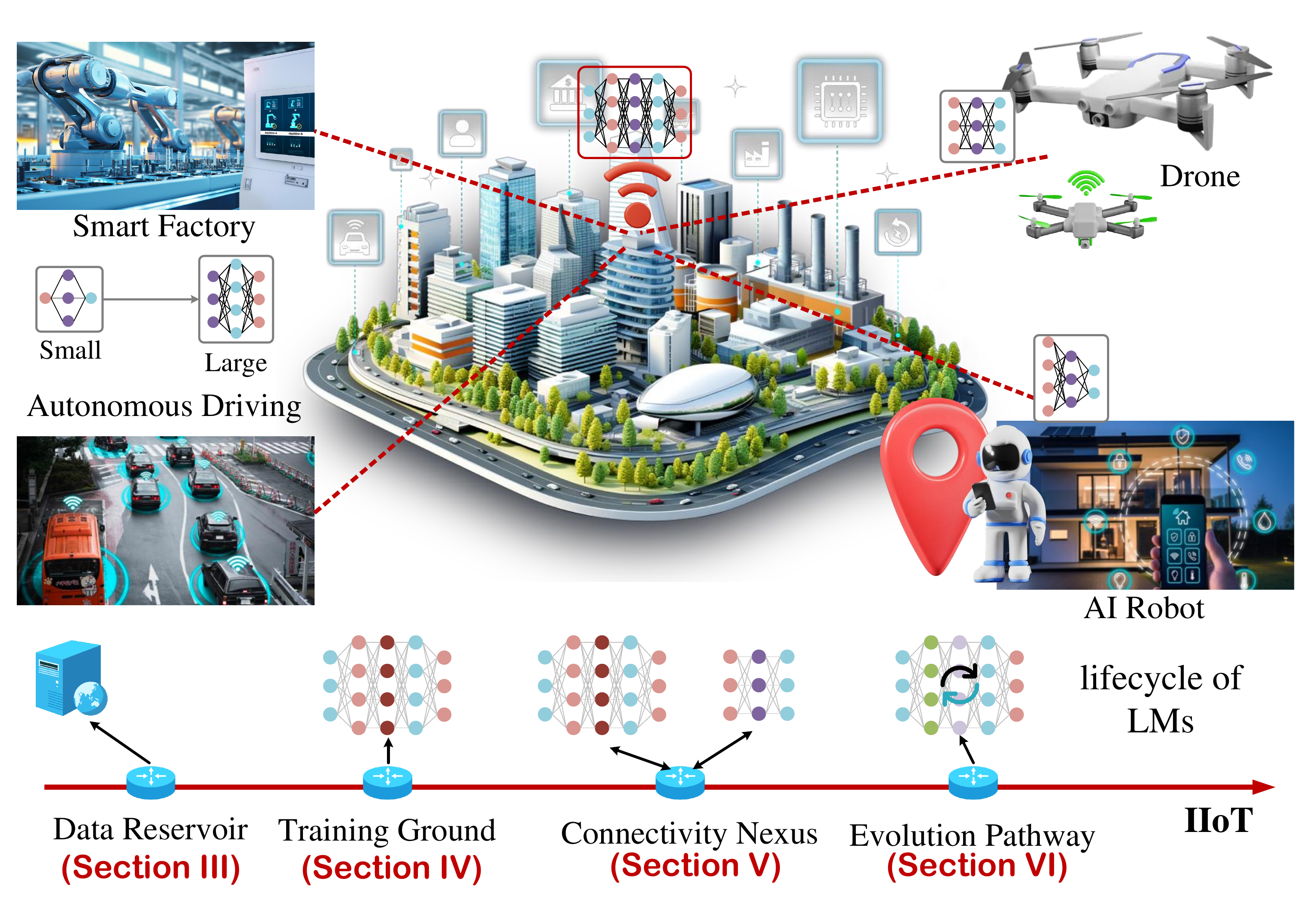}
    \caption{Industrial AI is transitioning from traditional deep learning models to large-scale transformer-based architectures, while the IIoT is also evolving in this process, advancing from a simple data pipeline to an intelligent infrastructure. \textbf{Our key insight} is that IIoT plays a critical role in supporting the full lifecycle of large models, including data foundation, model training, model connectivity, and continuous model evolution.}
    \label{fig:iiot}
\end{figure*}

IIoT serves as a critical framework for industrial digitalization, comprising three core components: sensing, computing, and networking and communication \cite{10018896,9036885,10697107,siam2024artificial}. The sensing layer gathers real-time data through sensors, cameras, and smart devices; the computing layer processes and analyzes data using edge and cloud computing; and the networking and communication layer enables efficient data transmission and inter-device connectivity with low latency and high bandwidth \cite{9885226,9963708,8792078}. This architecture not only supports traditional data collection and transmission but also evolves into an intelligent infrastructure, driving significant improvements in production efficiency and system intelligence \cite{8815938,9112632,9169769,9815106}.

As IIoT evolves rapidly, General Industrial Intelligence (GII) \cite{10697107,10440434,9790341,10102331} emerges as a frontier goal in the industrial domain. 
The architecture of GII is illustrated in Fig.~\ref{fig:gii} and comprises four layers: the IIoT infrastructure layer, the foundation layer composed of data and models, the algorithm layer, and the application layer.
GII aims to enhance the intelligence and flexibility of industrial systems while emphasizing human-centered design. Its core characteristics include:

\begin{itemize}
    \item Continual Learning and Optimization: GII systems continuously learn and optimize in dynamic environments to meet evolving production demands, surpassing traditional task-specific AI systems.
    \item Adaptation to Open Environments: GII systems operate effectively in complex and unstructured industrial settings, addressing the rapid changes and personalized demands of future manufacturing processes.
    \item Multimodal Interaction: GII systems interact with humans and the physical world through multimodal capabilities, including vision, language, and auditory modalities, significantly improving communication efficiency.
    \item Collaborative Intelligence: GII systems facilitate information sharing and knowledge collaboration among machines, production lines, factories, supply chains, and broader industrial ecosystems.
\end{itemize}
These characteristics underscore the role of GII in driving the deep integration of Industry 5.0 with cyber-physical-social systems \cite{ren2024aigc}.

The remarkable advancements in large models (LMs) represent a significant step toward achieving GII. Transformer-based models, including large language models (LLMs) \cite{vaswani2017attention, touvron2023llama, brown2020language}, vision models \cite{dosovitskiy2020image, radford2021learning}, time series models \cite{zhou2024one,wu2022timesnet}, and multimodal models \cite{liu2024visual, su2023pandagpt}, accumulate extensive knowledge through pretraining on large datasets, significantly enhancing the automation and diversity of data processing. For example, GPT-3 demonstrates success in language generation and reasoning tasks \cite{brown2020language}, while Vision Transformer achieves efficient performance in visual tasks \cite{dosovitskiy2020image}. Models like PandaGPT and LLaMA further expand multimodal interaction capabilities, enabling them to handle more complex tasks in industrial settings \cite{su2023pandagpt, touvron2023llama}. Despite their success, deploying these models in industrial scenarios requires addressing challenges such as dynamic data distributions, task complexity, and user-specific requirements.

\definecolor{blue1}{RGB}{229, 239, 247}  % Lightest blue
\definecolor{blue2}{RGB}{175, 205, 230}  % Lighter blue
\definecolor{blue3}{RGB}{135, 179, 213}  % Medium blue
\definecolor{blue4}{RGB}{95, 150, 191}   % Darker blue
\definecolor{blue5}{RGB}{50, 120, 170}   % Darkest blue
\begin{table*}[!t]
\centering
\caption{Survey comparisons on the lifecycle of large models with IIoT support.}
\label{tab:novel}
\resizebox{0.9\linewidth}{!}{
\begin{tabular}{lcccc}
\toprule
\textbf{Survey}                                                   & \textbf{Data Foundation}  & \textbf{Model Training}   & \textbf{Model Connectivity} & \textbf{Model Evolution}                                            \\ \midrule
IIoT Testbeds \cite{zhang2024survey}     & $\checkmark$                           &                           &    &                                                                                     \\ 
AIoT \cite{siam2024artificial}
& $\checkmark$  & $\checkmark$                           &                             &                                                                                     \\ 
Federated LLMs \cite{hu2024federated}            &  & $\checkmark$ &                             &                                                            \\ 
IoT and AIGC \cite{wang2024iot}               &                           & $\checkmark$  &    &                                                            \\ 
AIGC in MEN \cite{zhang2024toward}               &                           & $\checkmark$ &                             &                            \\ 
Agents Meet 6G \cite{xu2024large} &                           & $\checkmark$                           &    &                                \\ 
Embodied Intelligence \cite{10697107} &                           & $\checkmark$                           &    &                                \\ 
\rowcolor{blue2}
\textbf{Ours }                                                      & $\checkmark$ & $\checkmark$ & $\checkmark$   & $\checkmark$     \\ \bottomrule
\end{tabular}
}
\end{table*}

As illustrated in Fig.~\ref{fig:iiot}, \textbf{our key insight is that IIoT transitions from a traditional data pipeline to an intelligent infrastructure, playing a pivotal role in supporting the entire lifecycle of large models.} This transformation is reflected in the following aspects:

\begin{itemize}
    \item \textbf{Data Foundation}: IIoT serves as the primary source and repository for industrial data. Modern industrial systems generate high-frequency unimodal and multimodal data, including text, images, videos, code, and audio \cite{siam2024artificial}. We further explore how techniques such as data denoising, selection, and generation ensure data quality to support model training and fine-tuning.
    \item \textbf{Model Training}: IIoT plays a foundational role in model training by providing low-latency and high-bandwidth network infrastructure. This combination with edge computing effectively supports large-scale distributed training and real-time model fine-tuning, promoting efficient learning and dynamic adaptation.
    \item \textbf{Connectivity}: The deployment of large models in IIoT environments benefits from modular design and flexible routing mechanisms. These features significantly enhance system flexibility and adaptability, supporting diverse industrial needs and dynamic application scenarios.
    \item \textbf{Continuous Evolution}: IIoT enables large models to maintain long-term effectiveness and adaptability in dynamic industrial environments through continual learning mechanisms. By combining traditional continual learning methods with pre-trained model techniques, IIoT addresses data stream changes and incremental task updates, avoiding catastrophic forgetting and improving model performance.
\end{itemize}

In summary, IIoT evolves from a traditional data pipeline into an intelligent infrastructure, providing critical support across the entire lifecycle of large models. From data foundation to model training, connectivity, and continuous evolution, IIoT demonstrates its indispensable role. 
To ensure both relevance and quality, we focus on recent publications from the IEEE Communications Society and major AI conferences over the past five years.
This survey systematically analyzes the potential of integrating IIoT with large models, highlighting how this integration drives the transformation of industrial intelligence and offering theoretical frameworks and directions for future research.

\subsection{Distinguishing Features of This Survey}

Our innovations and main contributions are summarized as follows:

$\bullet$ \textbf{Systematic Analysis from a Full Lifecycle Perspective}
As shown in Table~\ref{tab:novel}, this paper is the first to systematically analyze the key role of IIoT in supporting large models from a full lifecycle perspective. Unlike other surveys that focus on individual stages (e.g., data foundation or model training), this paper comprehensively covers four core aspects: data foundation, model training, model connectivity, and continuous evolution. It reveals how IIoT provides comprehensive support for the development, deployment, and dynamic adaptation of large models.

$\bullet$ \textbf{Connectivity as a Driver of Emergent Intelligence}
This paper introduces the innovative concept of ``connectivity generates emergent intelligence” and explores the central role of IIoT in model connectivity. Through modular large model design, model routing, and model merging, IIoT enables single devices to support multi-task processing, such as dynamically allocating and combining multiple sub-models to handle complex tasks. This modularity and flexibility not only enhance the intelligence of industrial systems but also significantly expand the application potential of IIoT in dynamic environments, addressing the limitations of existing surveys that overlook the importance of connectivity.

$\bullet$ \textbf{Expanding the Industrial Applications of Continual Learning}
This paper deeply examines the critical role of continual learning mechanisms in dynamic industrial environments. It particularly focuses on how the combination of traditional continual learning methods with pre-trained model techniques enhances the long-term adaptability and effectiveness of models. Unlike existing studies that remain theoretical, this paper provides practical solutions tailored to the dynamic data streams and incremental task characteristics of IIoT, advancing the application of continual learning in industrial environments.

\section{Preliminary}\label{sec:Background}

\subsection{Timeline of GII}\label{sec:Research Roadmap}
\begin{figure*}[t]
    \centering
    \includegraphics[width=0.8\linewidth]{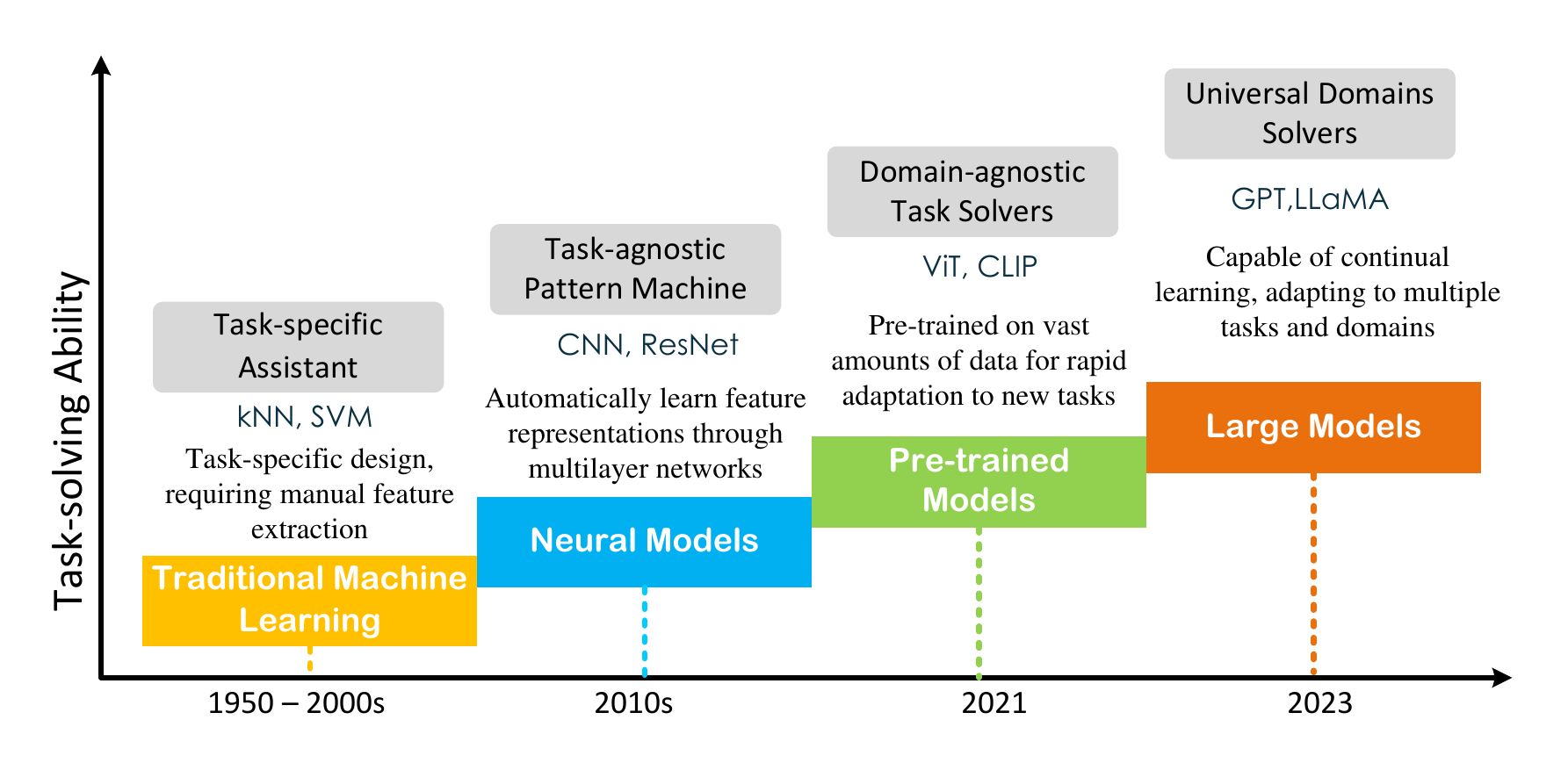}
    \caption{
    A roadmap for general industrial intelligence with four generations of models based on task solving capabilities.
    }
    \label{fig:Roadmap}
\end{figure*}

Illustrated in Fig.~\ref{fig:Roadmap}, GII has evolved through four crucial stages:

$\bullet$ \textbf{Traditional Machine Learning}, characterized by task-specific algorithm design optimized for small data sets, relying heavily on manual feature extraction and fine-tuning. Representative methods include kNN \cite{cover1967nearest} and SVM \cite{cortes1995support}.

$\bullet$ \textbf{Neural Networks}, marked by enhanced capabilities in handling large and complex data sets, with deep neural networks learning data features automatically, significantly improving pattern recognition accuracy and generalization. Representative models include CNN \cite{krizhevsky2017imagenet} and ResNet \cite{he2016deep}.

$\bullet$ \textbf{Pre-trained Models}, where models are pre-trained on massive data sets, enabling fine-tuning on relatively more minor specific tasks to achieve good performance, significantly reducing the time and resources needed for training task-specific models. Representative models include ViT \cite{dosovitskiy2020image} and CLIP \cite{radford2021learning}.

$\bullet$ \textbf{Large Models}, where models can continuously learn in changing tasks and environments, significantly expanding the boundaries of machine learning applications, marking a step towards higher levels of GII. Representative models include GPT \cite{achiam2023gpt} and LLaMA \cite{touvron2023llama}.

\subsection{Large Models}\label{sec:Industrial Large Models}
Table~\ref{tab:lms} summarizes the key large models for GII, highlighting their features and applications.
In Table~\ref{tab:lms}, the GPT series models are closed-source, meaning their parameters are neither accessible nor modifiable. Researchers can only guide the model's predictions by adjusting input data or employing techniques like prompt engineering. In contrast, the remaining models are open-source, allowing access to and modification of model parameters. This enables researchers to fine-tune the models for specific tasks, thereby significantly improving their performance.

\begin{table*}[!t]
\centering
\caption{Key Large Models for General Industrial Intelligence.}
\label{tab:lms}
\resizebox{0.9\linewidth}{!}{
\begin{tabular}{llll}
\toprule
\textbf{Model Type}                         & \textbf{Model Name}                                                                                  & \textbf{Key Features}                                                                             & \textbf{Applications}                                                                                                              \\ \midrule
\multirow{2}{*}{\textbf{LLMs}}              & \multirow{2}{*}{GPT, LLaMA}                                                                          & \multirow{2}{*}{\begin{tabular}[c]{@{}l@{}}Text generation\\ Semantic understanding\end{tabular}} & \multirow{2}{*}{\begin{tabular}[c]{@{}l@{}}Industrial fault diagnosis \\ Autonomous driving\end{tabular}}                          \\
                                            &                                                                                                      &                                                                                                   &                                                                                                                                    \\ \midrule
\textbf{Vision Models}                      & ViT, CLIP                                                                                            & \begin{tabular}[c]{@{}l@{}}Visual QA\\ Image classification\end{tabular}                          & \begin{tabular}[c]{@{}l@{}}Industrial anomaly detection \\ Human-robot collaboration\end{tabular}                                  \\ \midrule
\textbf{Time Series Models}                 & GPT4TS, Time-LLM                                                                                     & Time series specific                                                                              & \begin{tabular}[c]{@{}l@{}}Equipment predictive maintenance \\ Energy consumption forecasting\end{tabular}                         \\ \midrule
\multirow{2}{*}{\textbf{Multimodal Models}} & \multirow{2}{*}{\begin{tabular}[c]{@{}l@{}}PandaGPT, BLIP-2\\ LLaVA, LLaMA-Adapter\end{tabular}} & \multirow{2}{*}{Multimodal capabilities}                                               & \multirow{2}{*}{\begin{tabular}[c]{@{}l@{}}Industrial scene understanding\\ Industrial multimedia content generation\end{tabular}} \\
                                            &                                                                                                      &                                                                                                   &                                                                                                                                    \\ \bottomrule
\end{tabular}
}
\end{table*}

\subsubsection{Model Architectures}\label{sec:Model Architectures}
Transformer \cite{vaswani2017attention} is a deep learning model known for its proficiency in processing sequential data types like text or time series data.
Its workflow can generally be divided into several steps. 
Initially, input data is transformed into a set of embeddings, capturing initial data characteristics. 
Next, the embeddings go through a series of encoders and decoders. 
At each step, an attention mechanism enables the model to focus on different input data segments. 
Finally, the model generates output using auto-regressive decoding.
Specifically, a brief overview of the architectures in Transformer is provided below.

\textbf{Embedding.}
Embeddings convert input data (\textit{e.g.,} words, sentences, or any form of raw data) into fixed-length vector representations. 
Commonly used tokenizers, such as wordpieces and byte-pair encoding, are frequently employed in this process \cite{toraman2023impact}.
These vectors capture essential input features and are subsequently processed by the model.
In addition to embedding the input data, the Transformer also integrates positional encoding to maintain the sequential order of words in the sequence.
Vanilla Transformer employs Sinusoidal positional encoding \cite{vaswani2017attention} or Learned positional encoding \cite{gehring2017convolutional}.

\textbf{Attention Mechanism.}
The attention mechanism, which serves as the core of Transformer, enables the model to concentrate on various segments of the input sequence, thereby improving its comprehension of sequence relationships.
This process is achieved by calculating weight scores for each input position and using them to generate a weighted average output.
The most basic form of the attention mechanism (scaled dot-product attention) can be expressed by
\begin{equation}
    \text{Attention}(Q, K, V) = \text{softmax}\left(\frac{QK^T}{\sqrt{d_k}}\right)V,
\end{equation}
where $Q$, $K$, and $V$ represent the query, key, and value matrices, respectively, and $d_k$ denotes the dimension of the key. 
Transformer utilizes three distinct types of attention mechanisms, each focusing on different inputs:
i) Self-Attention: 
$Q$, $K$, and $V$ all come from the same sample.
ii) Encoder-Decoder Attention: The decoder allows focusing on the relevant positions in the input sequence. 
$Q$ is derived from the output of the prior decoder layer, while $K$ and $V$ come from the encoder's output. 
iii) Masked Self-Attention: The decoder's self-attention layers apply a mask to prevent attending to future positions, which is achieved by setting a part of the input to infinity before the self-attention softmax step.

\textbf{Multi-Head Attention.}
Multi-head attention enables the model to capture information from different representation subspaces simultaneously. 
In practice, this means passing $Q$, $K$, and $V$ through different linear transformations $h$ times, where $h$ is the number of heads, applying attention mechanisms separately, and subsequently concatenating the outputs. 
This allows the model to capture different aspects of the sequence at various positions, thereby enhancing the model's information-processing capabilities. 
Multi-Head Attention can be expressed by
\begin{equation}
    \text{MultiHead}(Q, K, V) = \text{Concat}(\text{head}_1, \dots, \text{head}_h)W^O,
\end{equation}
where each $\text{head}_i$ is
\begin{equation}
    \text{head}_i = \text{Attention}(QW^Q_i, KW^K_i, VW^V_i),
\end{equation}
where $W^Q_i$, $W^K_i$, and $W^V_i$ are parameter matrices for each head $i$, with $W^O$ as the output parameter matrix, and $h$ representing the number of heads.
By integrating these elements, the Transformer can process sequential data efficiently and effectively, capturing long-distance dependencies while maintaining relatively low computational complexity. This has made it the architecture of choice for various NLP tasks.

\textbf{Encoder-Decoder Architecture.}
In Transformer, the encoder and decoder are separate, each comprising multiple identical layers. 
Each layer in the encoder contains two sub-layers: a multi-head self-attention mechanism and a simple, position-wise, fully connected feed-forward network. 
Both sub-layers are equipped with residual connections, followed by layer normalization. 
The decoder closely resembles the encoder but incorporates two multi-head attention mechanisms instead of one. The first multi-head attention mechanism is masked to avoid positions from attending to the future.
An additional sub-layer is inserted between the encoder and decoder to pass the key and value information from the encoder to the decoder.

\textbf{Auto-regressive Decoding and KV Cache.}
Auto-regressive decoding generates model outputs sequentially, using previously generated outputs as input for the subsequent step. 
This method relies on earlier outputs to generate subsequent ones. 
In Transformer, a key-value cache \cite{dao2022flashattention} mechanism can enhance the efficiency of the decoding process by storing and reusing previous attention calculations, thus speeding up generation and increasing efficiency.

We introduce a data-centred taxonomy of industrial LMs. We classify existing LMs into four categories: language-based models, vision-based models, time series-based models, and multimodal models.

\subsubsection{Large Language Models}\label{sec:Large Language Models}
LLMs are predominantly constructed using the Transformer \cite{vaswani2017attention} architecture and encompass pre-trained LMs whose parameter counts span billions to trillions. These models, extensively pre-trained on vast textual corpora, have amassed a considerable depth of knowledge, empowering them to handle a variety of NLP tasks adeptly. Such tasks notably include text generation, semantic understanding, and sentiment analysis.

LLMs are categorized into open-source and closed-source categories.
Open-source models like LLaMA \cite{touvron2023llama}, Vicuna \cite{chiang2023vicuna}, and Gemma provide the research community and developers with flexibility by allowing direct access to the model architecture and pre-trained weights.
This access facilitates fine-tuning for specialized tasks, showcasing their robust performance in few-shot and zero-shot learning contexts. These contexts highlight their adaptability to new tasks using minimal or no new training data.

Conversely, closed-source LLMs, such as PaLM \cite{chowdhery2023palm} and GPT-4 \cite{achiam2023gpt}, are primarily available through API services. These services require users to formulate precise prompts to direct the models’ inference tasks. Although this modality restricts direct control over and transparency of the models' internal workings, it offers a user-friendly interface advantageous for commercial applications and product integration, enabling effective utilization of these powerful computational tools.

\subsubsection{Vision-based Models}\label{sec:Vision-based Models}
Early models in computer vision primarily used CNNs like ResNet \cite{he2016deep}, pre-trained on datasets such as ImageNet, to enhance performance on downstream tasks. The introduction of Vision Transformer (ViT) \cite{dosovitskiy2020image} marked a significant shift by applying the Transformer architecture to image patches, enabling the model to recognize image patterns without the traditional CNN inductive biases like local receptive fields and translational invariance. ViT depends heavily on extensive pre-training to extract discriminative features that capture both local and global image details.

Recent developments have introduced Vision Foundation Models (VFMs) that broadly fall into two categories: language-augmented and purely visual models. Language-augmented VFMs, like CLIP \cite{radford2021learning}, utilize contrastive learning to align text and image data, improving image encoding robustness. Purely visual VFMs, trained with techniques like masked autoencoders, focus solely on visual information, showing strong generalization in various vision tasks and notable zero-shot capabilities.

SAM \cite{kirillov2023segment} integrates image and prompt encoders with a mask decoder, all pre-trained on large datasets. It uses textual or spatial prompts to dynamically generate segmentation masks, enabling effective zero-shot adaptation across varying data distributions through prompt engineering. 

\subsubsection{Time Series-based Models}\label{sec:Time Series-based Models}
Initially applying LLMs to generic time series forecasting, a one-size-fits-all approach \cite{zhou2024one} introduces a unified framework using partially frozen LLMs.
Specifically, fine-tuning is limited to the embedding and normalization layers, while the self-attention and feedforward layers are frozen.
This strategy achieves state-of-the-art or comparable performance across major time series analysis tasks, including time series classification, short/long-term forecasting, interpolation, anomaly detection, and few-shot and zero-shot predictions. 

Recent work like Time-LLM \cite{jin2023time} proposes reprogramming time series with the source data modality and using natural language prompts to harness the potential of LLMs as effective time series models.
It achieves state-of-the-art performance across various prediction scenarios and excels in few-shot and zero-shot settings. 
Time-LLM is lightweight and efficient as it does not directly manipulate input time series data or fine-tune the underlying LLM.
In the industrial field, a method for fault diagnosis using time series data was proposed. This method converts 1-D time series signals into grayscale images using gray-transform and transform-gray coding methods for intuitive fault representation. 

\begin{figure}[t]
    \centering
    \includegraphics[width=0.95\linewidth]{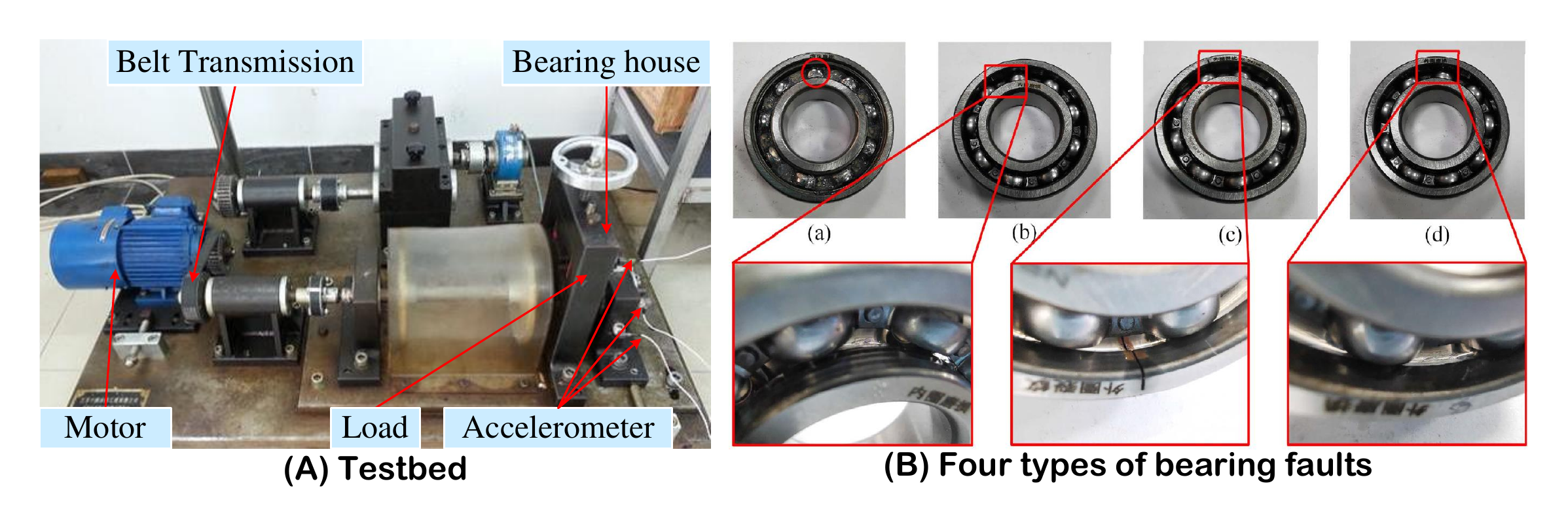}
    \caption{
    (A) Actual bearing fault testbed. (B) indicates four common types of bearing faults\textsuperscript{1}, \textit{i.e.}, (a) Rolling body wear, (b) Inner race wear, (c) Outer race crack, (d) Outer race wear.
    }
    \label{fig:fd}
\end{figure}
\footnotetext[1]{This image is referenced from literature \cite{yang2024novel}.}

\subsubsection{Multimodal Models}\label{sec:Multimodal Models}

Multimodal models are machine learning frameworks designed to process and analyze data from multiple input types, such as text, images, and audio, within a unified architecture. Unlike single-modality models that focus exclusively on one type of data (e.g., text-only or image-only), multimodal models integrate information across multiple modalities. This integration allows these models to leverage complementary features from different data types, resulting in enhanced task accuracy and improved robustness. For example, combining visual and textual information enables a deeper understanding of contextual relationships that would be missed by single-modality approaches.
Below are some notable examples:

$\bullet$ BLIP \cite{li2022blip} is a multimodal model focused on improving image description and comprehension tasks. By leveraging pre-training techniques like contrastive learning, BLIP effectively aligns textual and visual representations. Its successor, BLIP-2 \cite{li2023blip}, adopts a dual-stream Transformer architecture that processes visual and textual data streams independently before integrating them with an enhanced cross-modal attention mechanism. This approach significantly improves the alignment between images and text, enabling more accurate image description and understanding.

$\bullet$ MiniGPT-4 \cite{zhu2023minigpt} is a compact version of GPT tailored for multimodal tasks involving images and text. Despite its reduced scale, it maintains the generative capabilities of the GPT series while focusing on efficient processing of multimodal inputs, making it suitable for real-time applications.

$\bullet$ LLaMA-Adapter \cite{zhang2023llama} and LLaVA \cite{liu2024visual} employ adapter modules or simple linear layers to integrate visual features into language models. These methods aim to minimize interference between visual and textual data during training while optimizing the model’s ability to understand and generate visual information. This design enables efficient multimodal learning with reduced computational overhead.

\begin{table}[t]
\centering
\caption{The five core subtasks of fault diagnosis.}
\label{tab:fd_subtasks}
\begin{tabular}{ll}
\toprule
\textbf{Subtask}       & \textbf{Description}                                           \\ \midrule
Detection              & Determine the presence of a fault.                              \\
Classification         & Identify the type of fault present.                             \\
Severity               & Assess the severity of the fault.                               \\
RUL & Predict the time until maintenance is required.                \\
Fault tracing             & Investigate the underlying cause of the fault.                  \\ \bottomrule
\end{tabular}
\end{table}

\subsection{Key Tasks of Large Model in GII} \label{sec:Key Tasks of GII}
Building on the introduction to Large Models in Section II-B, this section focuses on the core industrial application capabilities enabled by these LMs. 
Industrial tasks are categorized into four primary types based on their nature and specific requirements: classification, detection/localization, planning \& control, and question \& answer (Q\&A). 
Each task type presents unique challenges and demands specialized technological approaches, particularly leveraging IIoT, edge computing, and advanced communication technologies. 
Detailed discussions on the applications and solutions tailored to each task type will be provided in the subsequent sections.

\subsubsection{Classification}\label{sec:Classification}
Classification tasks form the backbone of advanced analytics in industrial systems, especially when integrating IIoT data with LMs for enhanced measurement and decision-making accuracy. 
These tasks encompass multiple sub-domains, including industrial fault diagnosis and industrial human behavior recognition. LMs, with their robust data processing capabilities and advanced learning algorithms, significantly enhance the accuracy and efficiency of classification tasks. 

\textbf{Mechanical Fault Diagnosis.}\label{sec:Mechanical Fault Diagnosis}
The fault diagnosis tasks map measures signal features relative to the state of mechanical components \cite{10102331,10419797,chen2022globecom}. This process involves determining the basic condition of whether a fault exists. It may include more complex estimations such as the fault level, the severity of the fault, the remaining useful life, and the specific cause of the fault. Table~\ref{tab:fd_subtasks} details these five essential sub-tasks of fault diagnosis.
Equipment fault diagnosis demands high quality and quantity of data, especially when dealing with different types of equipment and their respective fault types, where relevant data are often scarce. Although there are some open-source datasets available, such as CWRU \cite{smith2015rolling}, and XJTU-SY \cite{wang2018hybrid}, which provide some support for model development, they still fail to fully meet the data requirements of practical applications. Fig.~\ref{fig:fd} illustrates a typical diagnostic data collection device capable of collecting essential operational data from equipment during actual operations.

Although LLMs such as ChatGPT \cite{achiam2023gpt} offer extensive knowledge bases and proficient language processing capabilities, their application in specialized domains like mechanical fault diagnosis is limited due to a lack of domain-specific expertise. Wang \textit{et al.} \cite{wang2023empowering} propose overcoming these constraints by integrating Local Knowledge Bases (LKB) with LLMs. This integration involves preparing the LKB, vectorizing its content, and applying prompt engineering to tailor the LLM's responses for domain-specific tasks.

\begin{figure}[t]
    \centering
    \includegraphics[width=0.98\linewidth]{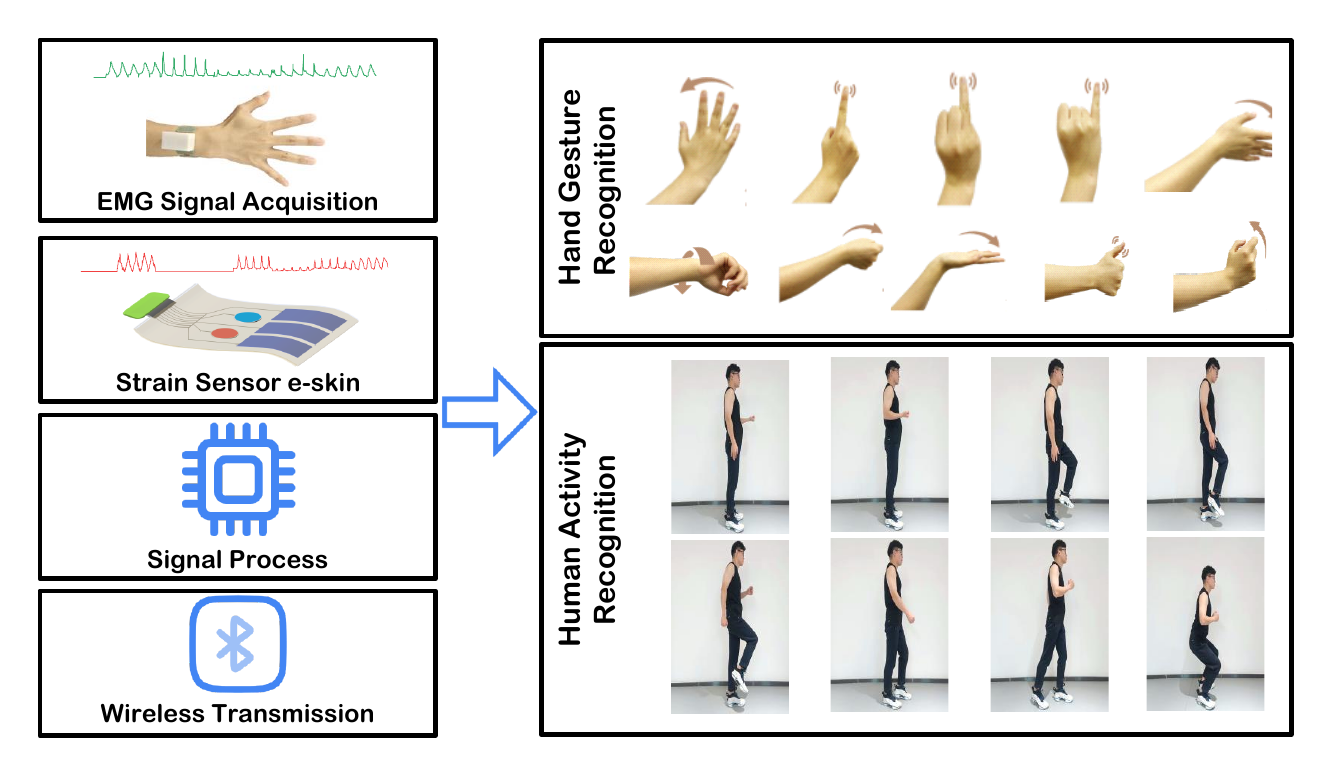}
    \caption{
    EMG/E-skin based human activity recognition and gesture recognition.
    }
    \label{fig:emg}
\end{figure}

\textbf{Activity/Gesture Recognition.}\label{sec:Industrial Human Action Recognition}
In automated systems, human activity, state monitoring, human pose estimation and gesture recognition technologies are crucial for quality control, safety monitoring, and process optimization. Typically, data is collected using flexible electronic skin or electromyography (EMG) sensors (see Fig.~\ref{fig:emg}). Flexible electronic skin detects changes in pressure and stress distribution on the skin surface, while EMG sensors monitor the electrical signals generated by muscles during activity. 

Flexible electronic skin, placed on key body parts, collects real-time physiological signals processed by deep learning models like CNNs for activity classification. Gesture recognition via surface electromyography involves attaching electrodes to the wrist, capturing, amplifying, filtering, and processing signals through deep learning to translate them into gestures, supporting human-computer interaction. Key components include electrodes, control boards, power units, and communication modules.

\subsubsection{Detection/Localization}\label{sec:Detection/Localization}
Detection and localization tasks primarily focus on identifying and precisely locating anomalies, defects, or specific features within industrial systems, which are crucial for ensuring product quality and process efficiency. LMs play a pivotal role in these tasks by enhancing the handling and analysis of measurement data. LMs facilitate sophisticated data preprocessing, enabling cleaner and more structured datasets that are essential for accurate detection. Moreover, their advanced capabilities in feature extraction allow for a deeper understanding of complex patterns within the data. Pattern recognition algorithms within LMs efficiently identify and categorize anomalies, while rigorous result validation processes ensure the reliability of the findings. 

\textbf{Industrial Human Action Recognition.}
In modern industrial settings, Industrial Human Action Recognition (IHAR) is crucial for enabling effective human-machine collaboration, improving quality control, safety monitoring, and optimizing processes. IIoT-connected cameras and depth sensors provide real-time contextual information critical for recognizing human actions in complex industrial environments. The data collected is processed locally at the edge, enabling immediate responses to safety-critical actions.

Recent advances in this domain leverage 3D-CNNs to interpret dynamic scenes and behaviors from video data. Additionally, techniques integrating Human-Object Interaction detection are increasingly applied to enhance the precision of action recognition. Despite these technological advancements, the application of machine learning in IHAR, especially in industrial environments, is still in its nascent stages. Liang \textit{et al.} \cite{liang2024low} have developed approaches that merge foundational models with lightweight architectures to support low-cost dataset creation and enable real-time IHAR recognition.

\begin{figure}[t]
    \centering
    \includegraphics[width=0.9\linewidth]{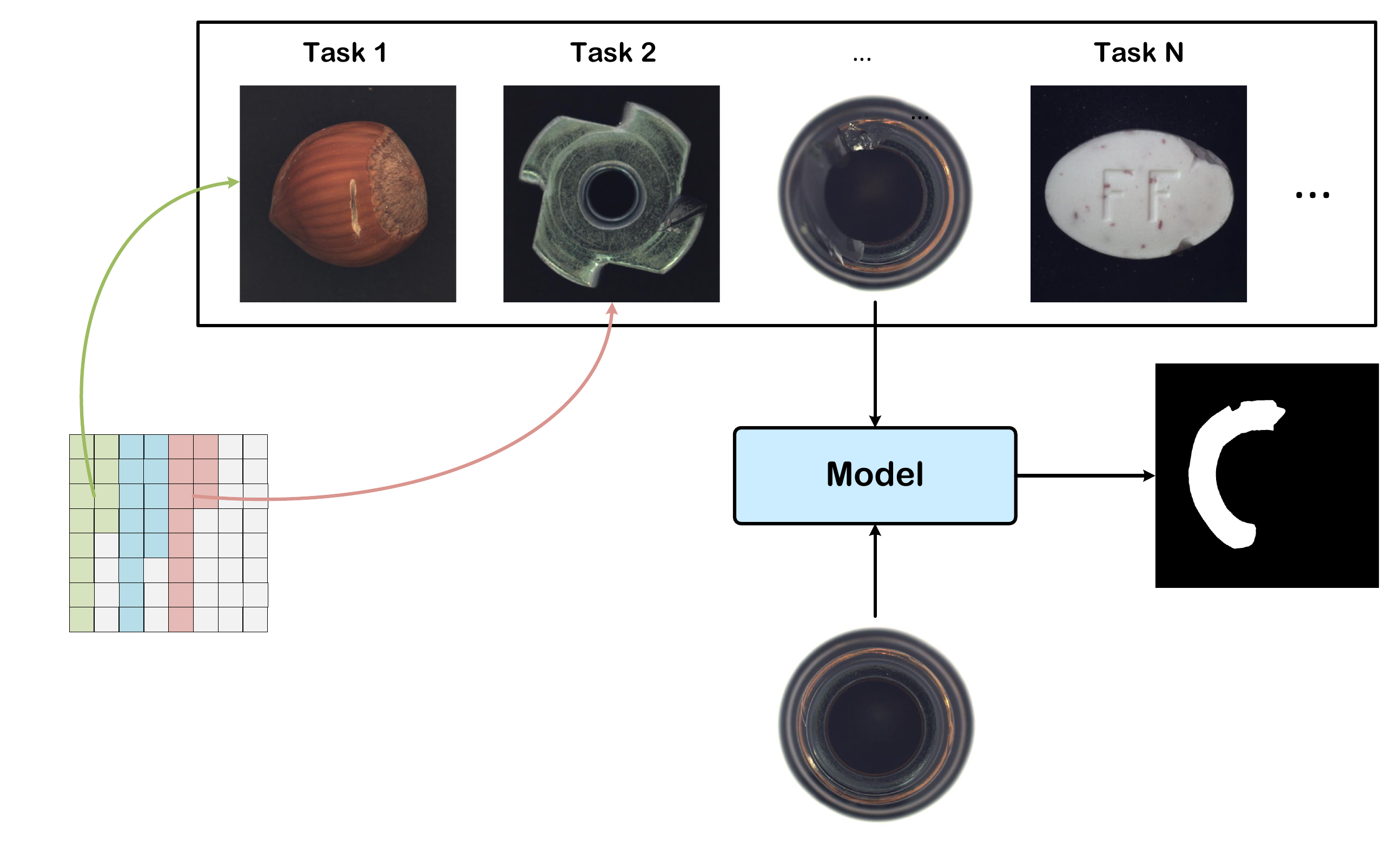}
    \caption{
    Industrial anomaly detection.
    }
    \label{fig:IAD}
\end{figure}

\textbf{Industrial Anomaly Detection.}\label{sec:Industrial Anomaly Detection}
As illustrated in Fig.~\ref{fig:IAD}, industrial anomaly detection aims to identify behaviours that deviate from normal operational patterns through continuous monitoring of production processes \cite{xie2024iad}. Popular datasets such as MVTec-AD \cite{bergmann2019mvtec} and VisA \cite{zou2022spot} are extensively used to benchmark detection models. MVTec-AD focuses on surface defect detection across various industrial objects with diverse materials and shapes, while VisA offers a broader range of visual anomalies, facilitating comprehensive evaluation of model generalization capabilities in complex scenarios. 

\begin{figure*}[t]
    \centering
    \includegraphics[width=0.8\linewidth]{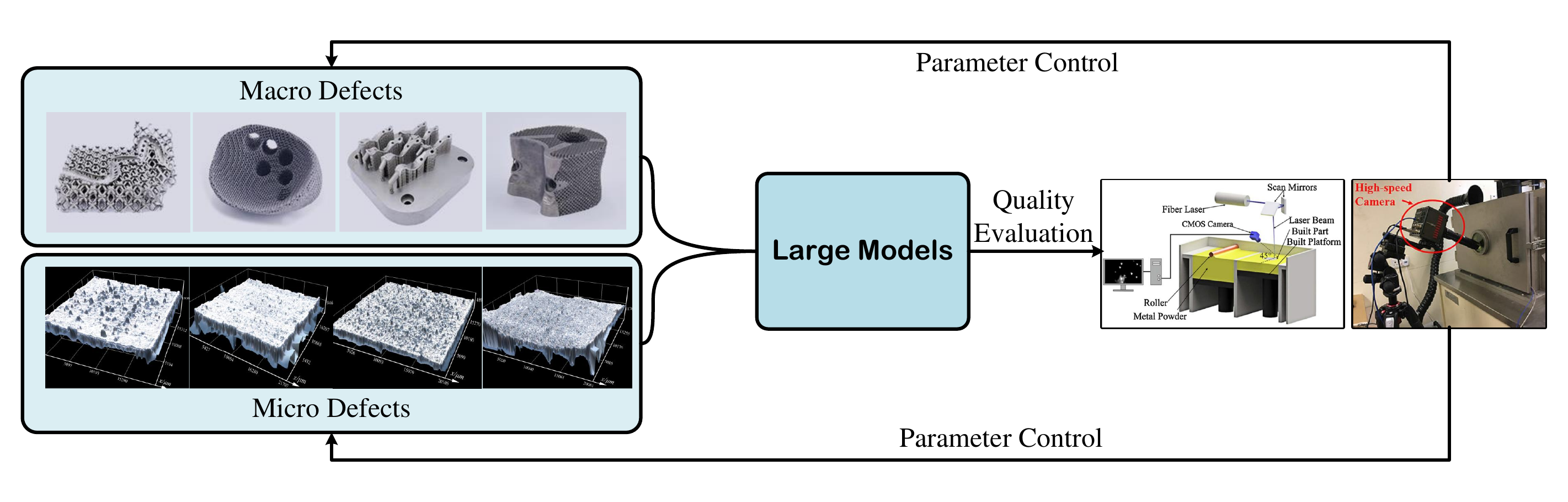}
    \caption{Additive manufacturing process monitoring.}
    \label{fig:LPBF}
\end{figure*}

\textbf{Laser Powder Bed Fusion.}\label{sec:Additive Manufacturing Process Monitoring}
Laser Powder Bed Fusion (LPBF) represents an advanced metal additive manufacturing technology, commonly referred to as 3D printing, as shown in Fig.~\ref{fig:LPBF}. This process utilizes a laser to selectively melt and solidify metal powder layer by layer, allowing for the fabrication of complex metal parts and the manufacturing of end-use components with intricate structures \cite{zhang2019powder}.

In LPBF, the deterioration of internal surface quality can compromise the mechanical performance of the final product and may even lead to processing failures. Ma \textit{et al.} \cite{ma2024layer} have designed an online monitoring system that adjusts monitoring parameters to address potential defects during the LPBF process, thus facilitating real-time control and quality improvement. The system incorporates a semantic segmentation model for visualizing surface irregularities and a ``surface defect area ratio" metric for quantifying defects.

In addition, Era \textit{et al.} \cite{era2023unsupervised} developed an image segmentation framework incorporating SAM \cite{kirillov2023segment} and a multi-point prompt generation scheme based on unsupervised clustering. This framework was applied to real-time porosity segmentation during the LPBF process and achieved high Dice Similarity Coefficients without the need for supervised fine-tuning of the model.

\textbf{Laser Sintered Processing.} 
Due to their superior mechanical, thermal, and dielectric properties, ceramic materials have become crucial in aerospace facilities, biomedical implants, and optoelectronic devices. Traditional ceramic processing methods face numerous challenges, particularly in material densification and dimensional accuracy. Laser sintering has emerged as a powerful tool for enhancing the fabrication of advanced ceramics, offering high resolution, rapid processing cycles, and flexible control \cite{lei2022direct}. 

\begin{figure}[t]
    \centering
    \includegraphics[width=0.9\linewidth]{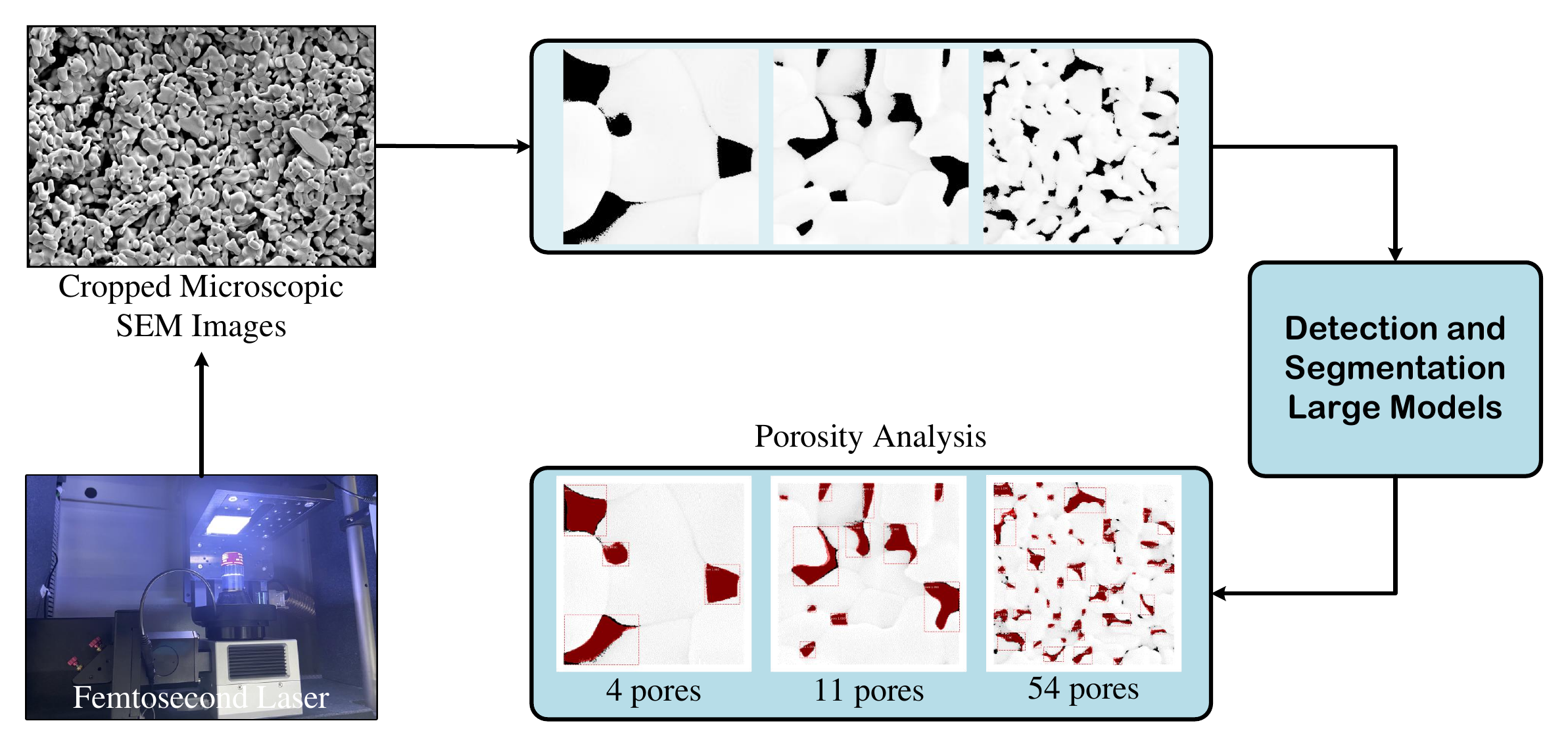}
    \caption{Porosity analysis for femtosecond laser processing.}
    \label{fig:Porosity_Analysis}
\end{figure}

However, the Gaussian distribution of laser energy leads to variations in material density and porosity during the sintering process, potentially reducing the finished products' mechanical strength and thermal stability. Therefore, accurate evaluation of the material property changes in laser-sintered materials is crucial for optimizing processing parameters to ensure the quality and performance of the end products. 
As shown in Fig.~\ref{fig:Porosity_Analysis}, Wang \textit{et al.} \cite{wang2023deep} have developed a micro-pore detection and segmentation model based on Mask R-CNN, providing a precise tool for measuring the surface porosity of laser-sintered alumina ceramics.

\subsubsection{Planning \& Control}\label{sec:Task Planning}
The Planning capabilities of LMs are essential for tasks like autonomous driving and robotics, where they optimize decision-making, tool learning \cite{qin2023tool}, route planning, and obstacle avoidance using real-time sensor data. This enables efficient navigation and task execution in dynamic environments. On the Control side, LMs play a key role in communication-control integration, optimizing signal processing, managing network resources, and ensuring reliable data transmission. This is crucial for real-time coordination between edge devices and centralized systems, enhancing both planning and control functions with minimal latency and high accuracy.

\begin{figure*}[!t]
    \centering
    \includegraphics[width=0.75\linewidth]{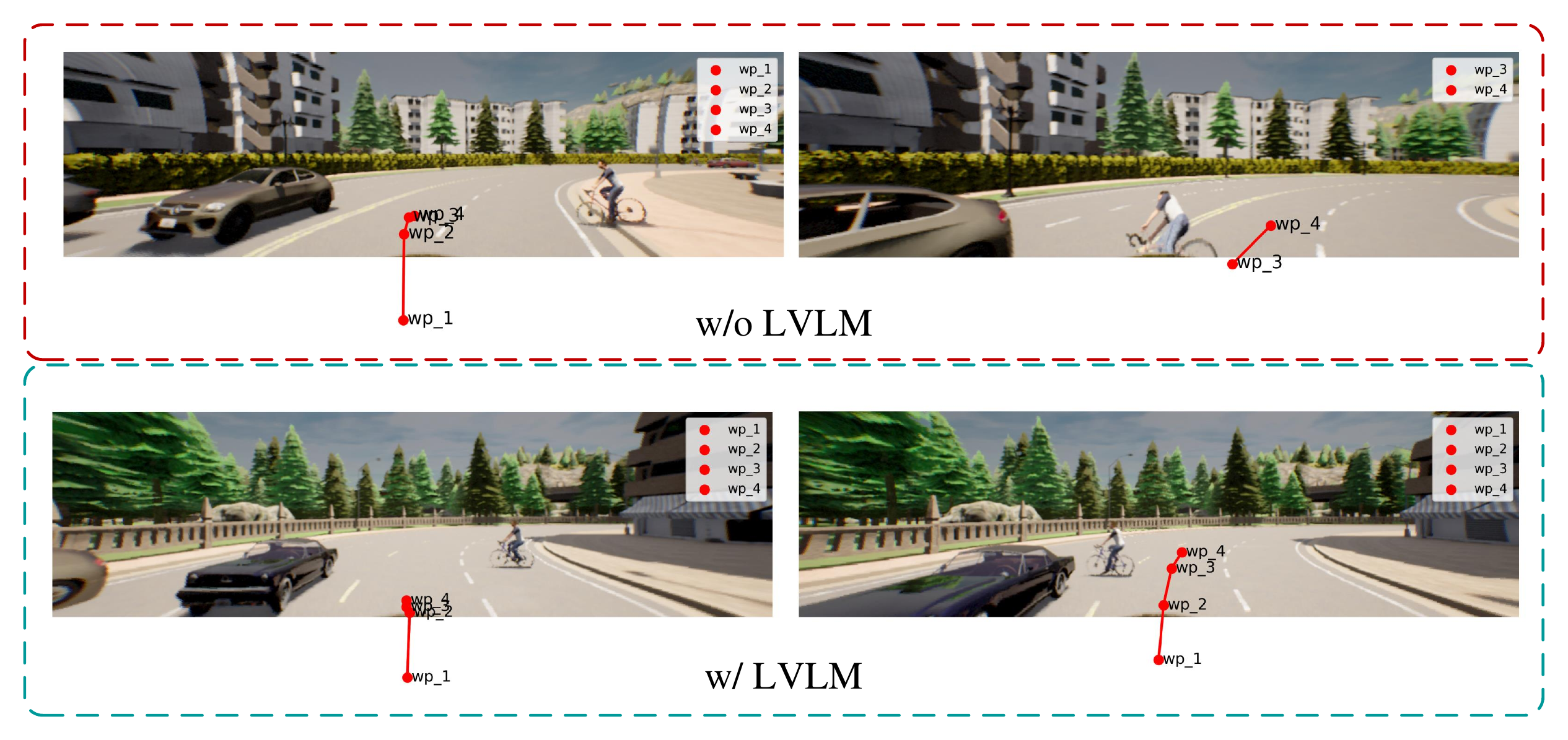}
    \caption{LLM planning and control for autonomous driving.}
    \label{fig:driving}
\end{figure*}

\textbf{Autonomous Driving.}\label{sec:Autonomous Driving Perception and Motion Planning}
Autonomous driving services depends on active sensing from modules such as camera, LiDAR, radar, and communication units \cite{10213996}.
As shown in Fig.~\ref{fig:driving}, integrating LLMs with the IIoT offers great potential for enhancing vehicle personalization and adaptability in autonomous driving, particularly in open-world scenarios \cite{10804108, chen2024edge}.
The process begins with speech-to-text technology converting verbal descriptions into text, which LLMs then interpret to generate executable path planning code. This code is tested in simulation environments, providing feedback for optimizing vehicle operation, enhancing adaptability, and decision-making accuracy \cite{sima2023drivelm, han2024dme}.

In dataset developments, the Honda Research Institute's HAD dataset  includes over 5,675 annotated driving video clips \cite{ma2023dolphins}, while the NuPrompt \cite{wu2023language} expands the nuScenes dataset with detailed object-centric linguistic annotations suitable for complex driving scenarios. The NuScenes-QA \cite{chen2023driving} provides extensive visual scene data for testing visual question-answering capabilities in autonomous driving applications.
DriveLM \cite{sima2023drivelm}, utilizing datasets like nuScenes and CARLA, showcases effective methods for combining graphical visual question answering with end-to-end driving tasks, proving the value of integrating visual reasoning in autonomous vehicle.

Fig.~\ref{fig:driving} illustrates the application of large models in autonomous driving scenarios. The first row depicts the situation without a large model, where the ego vehicle, despite decelerating, maintains a relatively high speed when encountering a pedestrian crossing, ultimately resulting in a collision. In contrast, the second row demonstrates the scenario when a ``stop" command is issued. The ego vehicle promptly executes a braking response, reducing its speed from 10.65 m/s to 1.06 m/s within just 2 seconds, successfully avoiding the collision.

\textbf{Industrial Robotics.}\label{sec:Industrial Robotics}
In recent advancements, researchers have focused on the integration of LLMs to enhance robotic operations in industrial settings. 
Fan \textit{et al.}\cite{fan2024embodied} explore the application of LLMs in the generation, error analysis, and code production of robotic tasks within manufacturing settings. 

\textbf{Industrial Control.}\label{sec:Industrial Control}
Industrial Control Systems control and operate the physical processes of critical national infrastructure sectors, such as water treatment, electricity generation, and manufacturing \cite{10100622}.
Industrial control tasks manage, regulate, and direct systems, equipment, or processes to achieve stable operation and the desired behaviour of the system. Song \textit{et al.} \cite{song2023pre} demonstrated the potential of foundational models in industrial control by employing clustering methods to select exemplars and generate prompts for a base model, which then regulated the air conditioning in buildings accordingly.
Joglekar \textit{et al.} \cite{joglekar2024towards} propose a global control strategy based on LLM that allows the transfer of control strategies to a limited set of skills. These skills can execute high-precision tasks through dynamic context switching when specifically trained.

\textbf{Co-Design of Communication and Control.}\label{sec:Co-design}
As a critical aspect of building the IIoT, the design of communication and control systems has always been a major focus\cite{ref1}. Due to the strong coupling relationship between the two, it is crucial to develop new metrics that can simultaneously evaluate the performance of both communication and control systems \cite{ref2}. 
Wang \textit{et al.} \cite{ref3} used the Age of Information (AoI) to characterize ``information freshness" and studied the impact of average AoI on control costs. By minimizing AoI, they aimed to reduce the combined costs of control and communication. Chen \textit{et al.} \cite{ref4,chen2024AoLI} considered scenarios with multiple users, addressing both uplink and downlink wireless transmissions. Their work integrated multiple antennas on the remote end to improve signal reception quality and developed a loop information age metric. This metric not only captures information freshness but also dynamically allocates the rounds of transmission attempts based on the success or failure of UL and DL transmissions, thereby enhancing communication reliability while ensuring control convergence speed.

\begin{figure}[t]
    \centering
    \includegraphics[width=1\linewidth]{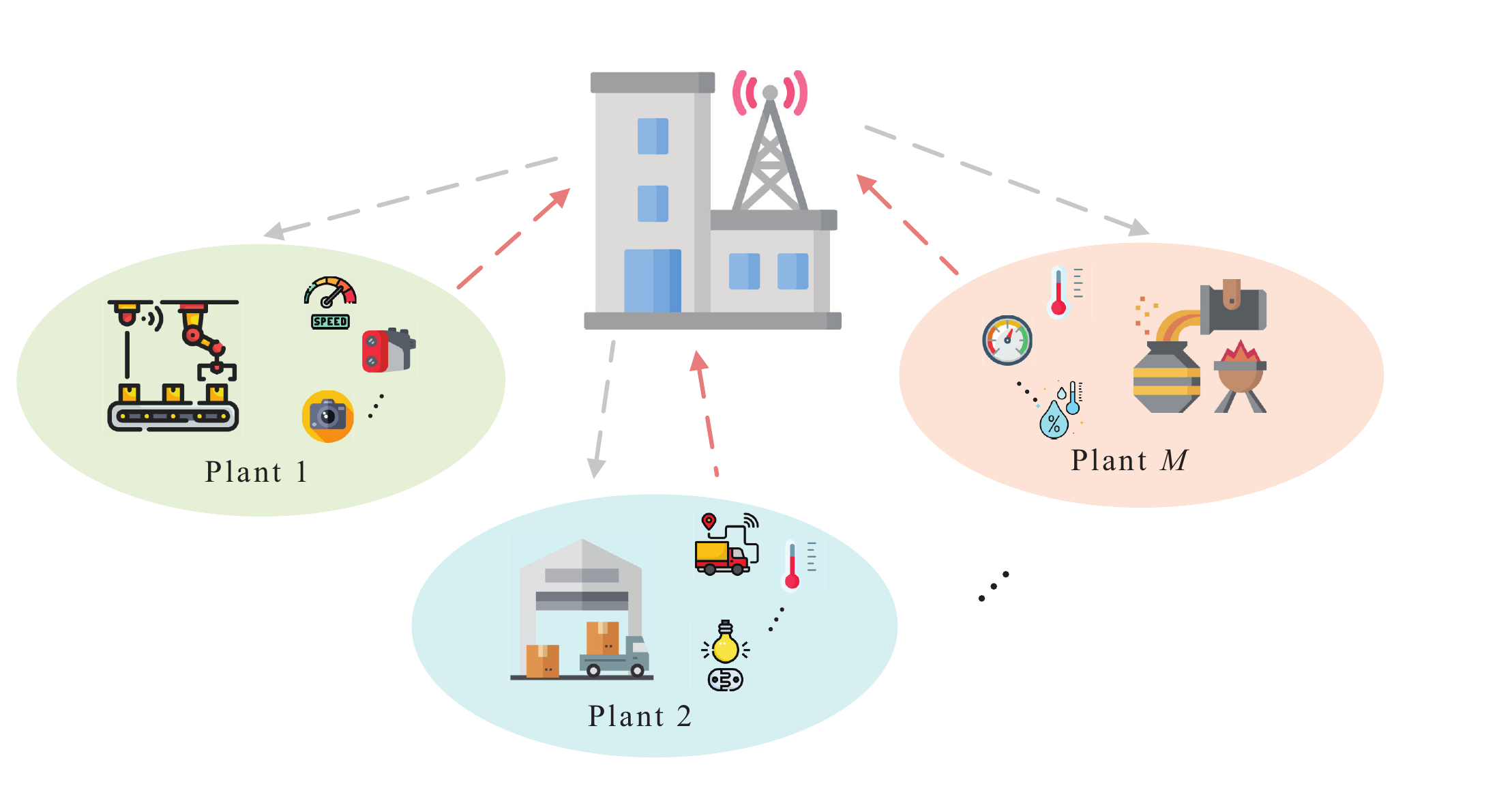}
    \caption{
    Communication and control system \cite{chen2024AoLI}.
    }
    \label{fig:Communication_control}
\end{figure}

As shown in Fig.~\ref{fig:Communication_control}, the co-design of communication and control enhances the stability and responsiveness of IIoT systems by optimizing data transmission and control. LMs, with their strong representation capabilities, offer significant potential in this area. By extracting semantic information \cite{zhang2022deep}, LMs can optimize system coordination and improve both communication efficiency and control response, paving the way for more advanced semantic communication and the further development of GII.

\subsubsection{General Q\&A}\label{sec:General QA}
\begin{figure}[t]
    \centering
    \includegraphics[width=1\linewidth]{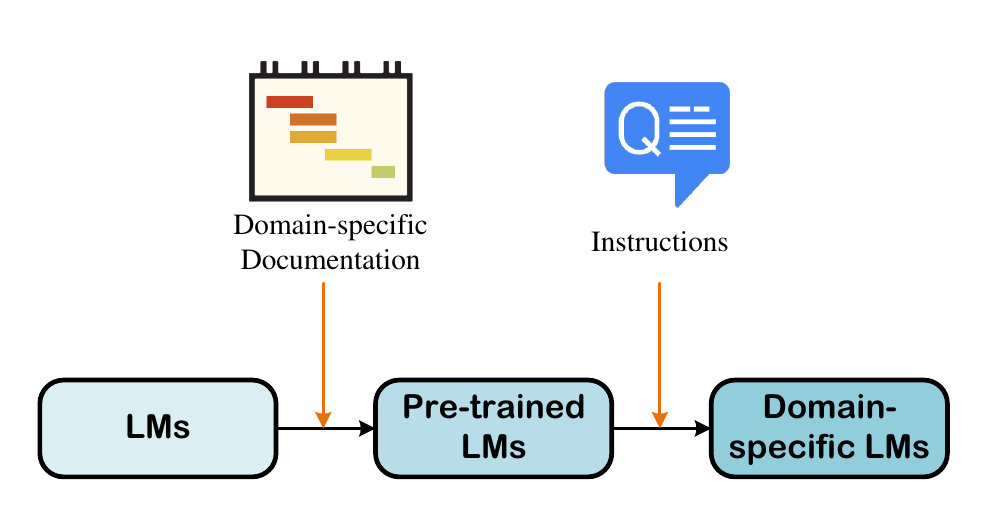}
    \caption{
    LLMs for industrial question answering.
    }
    \label{fig:industrial_qa}
\end{figure}
In the industrial sector, QA systems play a crucial role in addressing complex queries, supporting decision-making, and improving operational efficiency. As illustrated in Fig.~\ref{fig:industrial_qa}, these systems integrate into enterprise knowledge management frameworks to assist engineers and operators in quickly finding solutions, thereby reducing downtime and optimizing maintenance processes.

While LLMs perform well in open-domain QA tasks, their application in specific industrial domains presents unique challenges. Retrieval-based LLMs often struggle with complex industrial queries requiring deep domain-specific knowledge. Strategies to address these limitations include novel interaction paradigms that incorporate domain documentation and fine-tuning techniques to provide relevant knowledge during runtime. Furthermore, specialized QA datasets for industrial applications, such as the 32k QA pair dataset in the cloud domain by Wang \textit{et al.} \cite{wang2023empower}, establish new benchmarks for evaluating and improving LLM performance in industrial contexts.

\subsection{LMs Evaluation}
To effectively evaluate the performance of LLMs, a common approach is to select different capability dimensions and construct corresponding evaluation tasks, which are then used to test and compare the model's performance. The capability dimensions include language generation, knowledge utilization, and complex reasoning, among others. Depending on the evaluation method, the assessment approaches for these capability dimensions can be classified into three categories: benchmark-based methods \cite{hendrycksmeasuring}, human evaluation-based methods \cite{zheng2023judging}, and model evaluation-based methods. Table~\ref{tab:evaluation} lists typical works associated with each evaluation method.

\begin{table}[!t]
\centering
\caption{Typical Evaluation Work.}
\label{tab:evaluation}
\resizebox{1\linewidth}{!}{
\begin{tabular}{lllll}
\toprule
\textbf{Evaluation Work} & \textbf{Capability}   & \textbf{Data Source} \\ \midrule
MMLU \cite{hendrycksmeasuring}                     & General               & Human Exams          \\
BIG-Bench \cite{srivastava2022beyond}                & General               & Human Annotated      \\
C-Eval \cite{huang2024c}                   & General               & Human Exams          \\
Chatbot Arena \cite{zheng2023judging}            & Human Interaction     & Human Annotated      \\
AlpacaEval \cite{li2023alpacaeval}               & Instruction Following & Synthetic            \\
MT-Bench \cite{zheng2023judging}                 & Human Interaction     & Human Annotated      \\ \bottomrule
\end{tabular}
}
\end{table}

\section{IIoT: Data Reservoir for LMs}
\subsection{Sensing}
Sensing technologies are fundamental to the realization of intelligent IIoT systems, enabling task recognition, environmental perception, and system optimization through diverse data modalities \cite{8419202}. The performance metrics, evaluation indicators and architecture of data collection, data routing, virtualization, and other aspects have gained widespread attention \cite{9740691}.

\begin{itemize}
    \item \textbf{Motion Data}  
    Inertial Measurement Unit (IMU) sensors, including accelerometers, gyroscopes, and magnetometers, are widely used for capturing motion data.  
    Liu et al. \cite{liu2016lasagna} proposed the Lasagna framework, an unsupervised learning-based Human Activity Recognition (HAR) method. This framework extracts shared fundamental features of human motion in an unsupervised manner, building multi-resolution general representations for common human activities, significantly enhancing the expressiveness and utility of motion data.

    \item \textbf{Wireless Signals}  
    Wireless communication technologies, such as RFID, Wi-Fi, millimeter-wave, LTE, and LoRa, provide valuable signal data for various industrial applications.  
    Li et al. \cite{li2016deep} designed an RFID-based sensing system by attaching RFID tags to objects in clinical environments. They recorded the received signal strength from these tags and used the collected data as input to a CNN. This method enabled activity recognition involving specific objects, demonstrating significant potential in healthcare scenarios.

    \item \textbf{Vision Information}  
    Vision data is captured using RGB cameras, depth cameras, and Near-Infrared (NIR) image sensors.  
    Shim et al. \cite{shim2023mosaic} employed NIR image sensors to monitor human activities. Unlike RGB-based vision systems, NIR images inherently lack sufficient information to reveal personal identities, effectively addressing privacy concerns in applications requiring privacy-sensitive monitoring.

    \item \textbf{Acoustic Signals}  
    Acoustic sensors capture sound-based data for various industrial scenarios.  
    Mao et al. \cite{mao2020deeprange} introduced the DeepRange system, which addressed the limitations of traditional signal processing methods in utilizing aquatic signals for localization tasks, particularly in low Signal-to-Noise Ratio environments. They demonstrated that Deep Neural Networks could autonomously learn features from received acoustic signals to estimate distances, outperforming traditional signal processing algorithms in certain scenarios.

    \item \textbf{Point Cloud Data}  
    LiDAR and other 3D sensors generate high-precision point cloud data for advanced applications.  
    Point cloud data has garnered significant attention in end-to-end autonomous driving systems \cite{zhang2024deflow}. For instance, high-precision point cloud data generated by LiDAR provides detailed three-dimensional environmental information, supporting reliable path planning, object detection, and environmental perception in autonomous vehicles.

\end{itemize}

\subsection{Data Freshness}
In the IIoT environment, combining Age of Information (AoI) and Age of Incorrect Information (AoII) metrics, intelligent scheduling strategies, multi-access protocols, and sleep scheduling mechanisms effectively ensures the freshness and accuracy of information, thereby enhancing the efficiency and reliability of monitoring and control systems \cite{9377627}.

\begin{itemize}
    \item \textbf{Assessing Information Freshness:} 
    AoI serves as a critical metric for quantifying freshness in state update systems by measuring the time elapsed since the destination node last received updated information \cite{9380899,9681851}. Unlike traditional communication delay metrics, AoI comprehensively accounts for transmission, propagation, and queuing delays. To further evaluate both freshness and accuracy, AoII \cite{10638762} extends AoI by considering whether the information at the remote server aligns with the source's actual state, providing a more holistic perspective.

    \item \textbf{Intelligent Scheduling Strategies:} 
    Traditional scheduling methods, such as simple priority queues or predefined rules, often fail to address the complexity and dynamism of IIoT environments. Intelligent scheduling, particularly those leveraging DRL, dynamically optimizes scheduling by learning from environmental interactions \cite{10638762}. DRL excels in handling large state and action spaces while adapting to the heterogeneity of source updates and bandwidth constraints in IIoT.

    \item \textbf{Multi-Access Protocols and Sleep Scheduling:} 
    To reduce power consumption, various sleep scheduling strategies, such as Multi-Vacation and Start-up Threshold strategies, are applied in IIoT systems, allowing devices to enter sleep mode when idle \cite{9681851}. When combined with multi-access protocols like TDMA, FDMA, and NOMA, these strategies further enhance resource allocation and network performance. Different combinations of protocols and strategies exhibit varying impacts on the system's AoI and AoII performance.
\end{itemize}

\subsection{Data Augmentation}
\begin{figure}[t]
    \centering
    \includegraphics[width=0.9\linewidth]{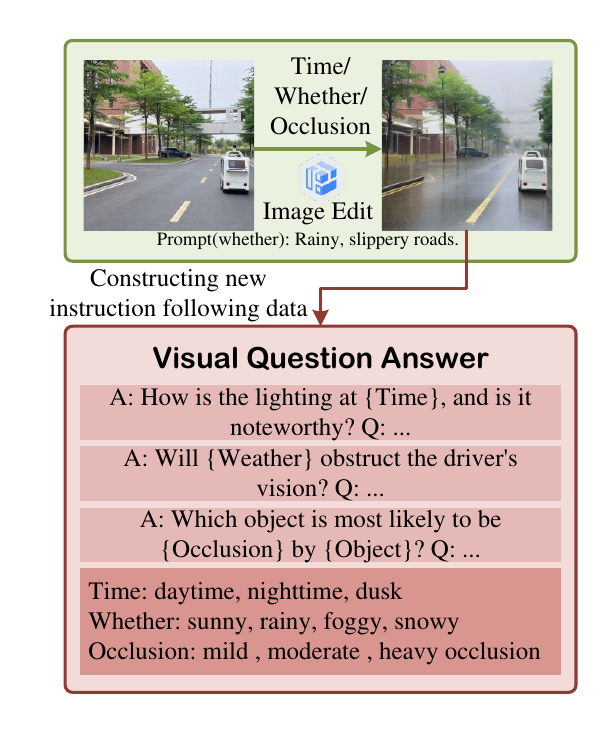}
    \caption{
    Data augmentation.
    }
    \label{fig:feature_diversity}
\end{figure}
Data augmentation plays a critical role in enhancing the performance and robustness of LLMs by enriching the training data with synthesized or modified examples.

For instance, LLMs are particularly adept at augmenting data interpretability by generating textual descriptions for non-textual data \cite{liu2023biosignal}. 
LLMs also excel in integrating diverse industrial data sources, thereby enhancing the robustness of data processing and decision-making capabilities. For example, Hou \textit{et al.} \cite{yu2023temporal} utilize GPT-4 to refine hierarchical feature definitions, significantly reducing the reliance on manual annotations and improving model stability during testing phases.
As shown in Fig.~\ref{fig:feature_diversity}, Chen \textit{et al.} enhance original images using an image-editing large model across three dimensions—time, weather, and occlusion \cite{10804108}. 

The following sections first outline the strategies employed by LLMs for data denoising and selection.

\subsubsection{Data Denoise}\label{sec:Data Denoise}
Acquiring large-scale datasets with precise expert annotations in practical settings is costly in terms of time and human resources. Consequently, real-world data, which inherently includes label noise, is commonly used. Addressing this challenge, Cheng \textit{et al.} \cite{cheng2023adaptive} propose an adaptive label noise learning framework based on pre-trained models. This framework integrates an adaptive warm-up phase, sample correctness statistics-driven data separation, and a linear decay fusion strategy to handle diverse label noise effectively. Unlike traditional methods targeting specific noise types, this approach enhances learning from noisy data, improving the accuracy and robustness of deep neural networks in real-world applications.

\subsubsection{Data Selection}\label{sec:Data Selection}
Data selection involves identifying subsets of available data that notably enhance model performance. Li \textit{et al.} \cite{li2023one} propose a quality-based mechanism to evaluate the influence of data on model perplexity, with an emphasis on prioritizing data that demonstrates high utility and training potential. In addition to this, techniques such as data pruning, active learning, and transfer learning are also employed to optimize datasets, all aimed at improving learning efficiency and overall outcomes.
In industrial applications, such as robotics \cite{10235949} and anomaly detection \cite{gu2024anomalygpt}, selecting high-quality data is crucial for enhancing operational efficiency and the accuracy of decision-making processes. This adaptive data selection approach supports continuous learning in dynamic environments, allowing industrial systems to maintain both stability and efficiency in the face of production shifts and external influences.
Moreover, advanced data selection strategies, such as the importance re-sampling method utilized by Xie \textit{et al.} \cite{xie2024data}, further optimize LLMs training by focusing on the most informative samples. This prioritization leads to substantial improvements in training efficiency and contributes to enhanced model performance.

\subsubsection{Reprogramming}\label{sec:Model Reprogramming}
\begin{figure}[t]
    \centering
    \includegraphics[width=1\linewidth]{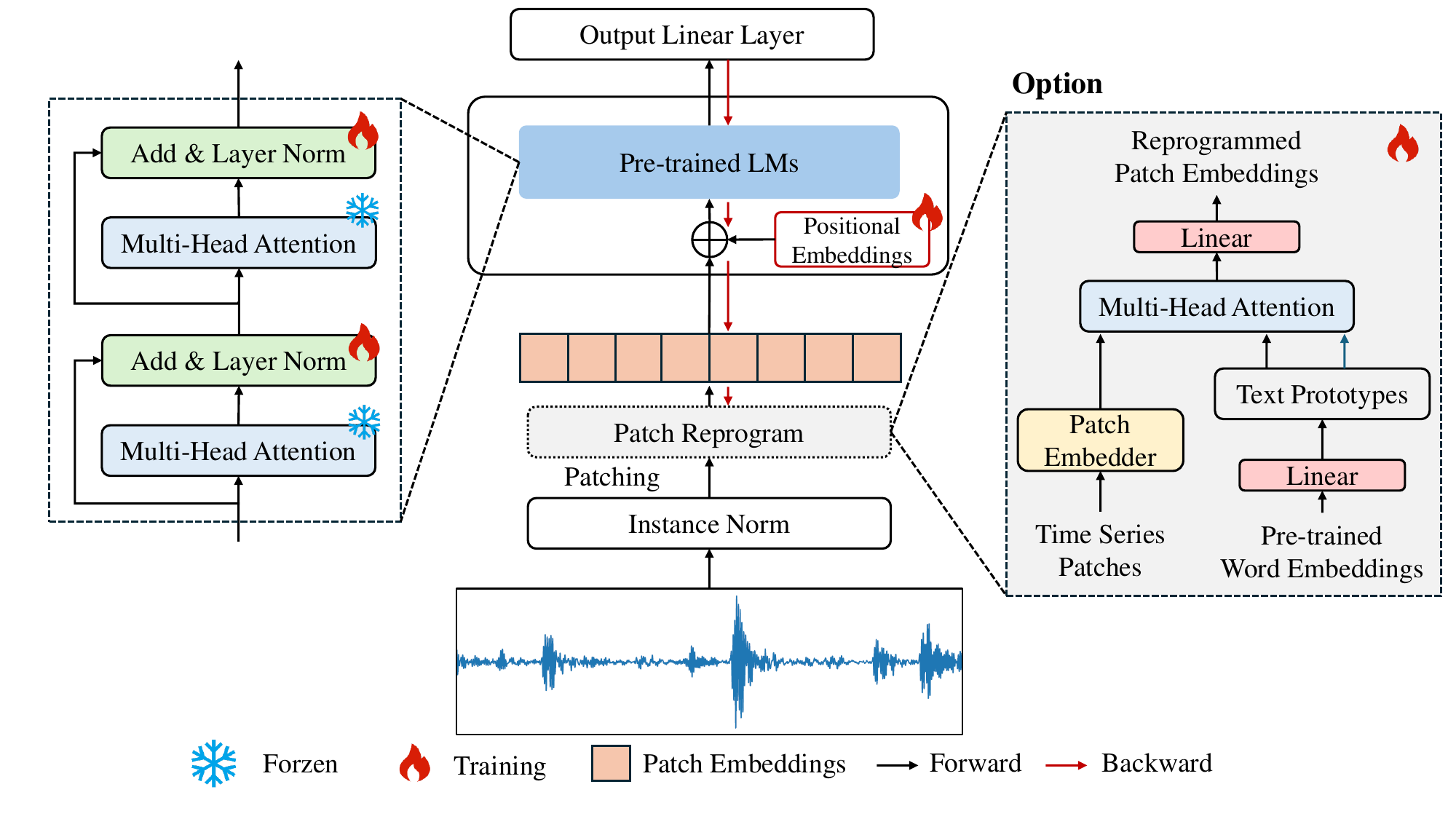}
    \caption{
    Model reprogramming.
    }
    \label{fig:Model_Reprogramming}
\end{figure}
Due to various challenges faced by deep learning in resource-constrained domains, such as limited data availability, constrained model development costs, and insufficient pre-trained models for effective fine-tuning, Chen \cite{chen2024model} proposes model reprogramming as a solution to bridge this gap.
As shown in Fig.~\ref{fig:Model_Reprogramming}, model reprogramming enables resource-efficient cross-domain machine learning by repurposing pre-trained models from a source domain to solve tasks in a target domain without fine-tuning, even in potentially significant differences between the domains.
In many applications, model reprogramming surpasses transfer learning and training from scratch in performance.

\subsection{Data Generation}\label{sec:Data Generation}

\begin{figure}[t]
    \centering
    \includegraphics[width=0.9\linewidth]{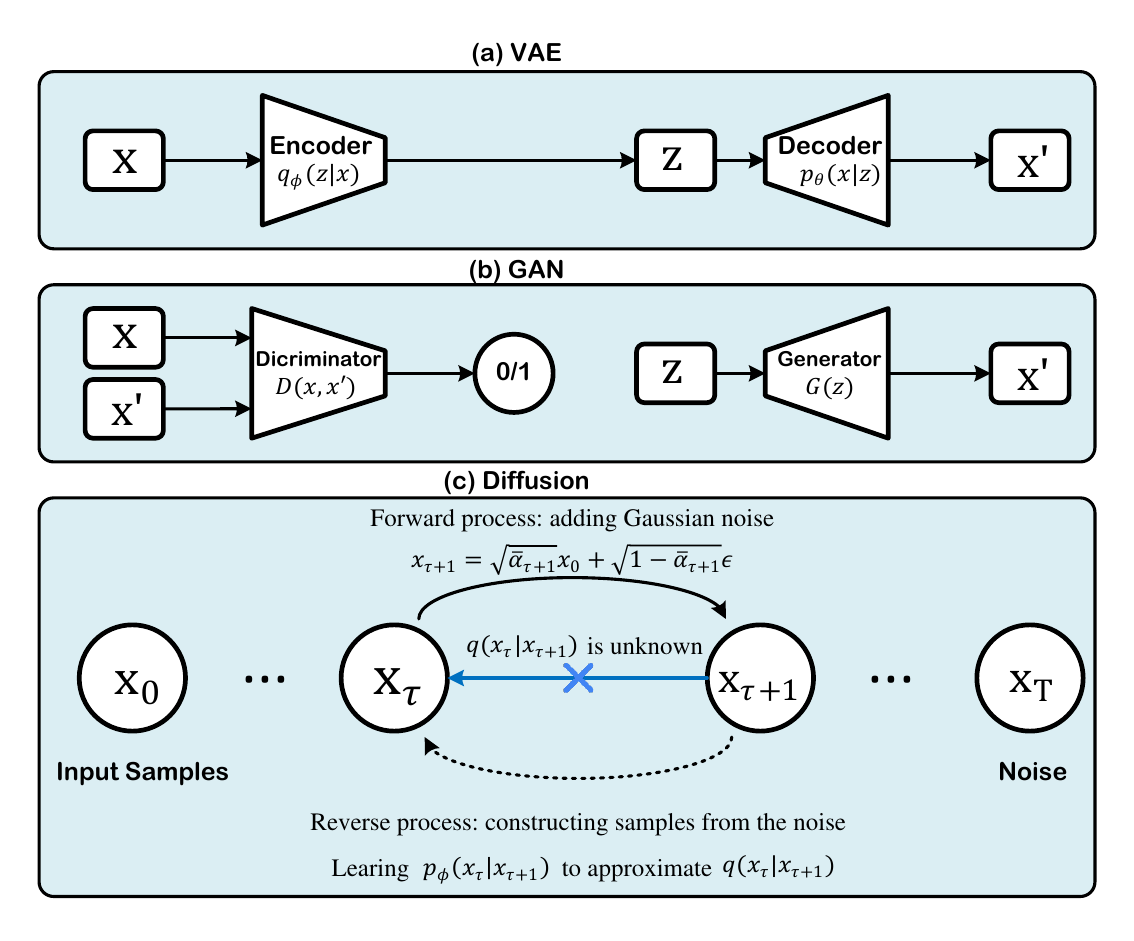}
    \caption{
    Different types of generative models.
    }
    \label{fig:generator}
\end{figure}

\subsubsection{Generative Models}
Deep learning tasks, traditionally supervised, heavily rely on costly and time-consuming labeled data annotation. To address this, current research explores zero-shot learning through generative dataset creation, leveraging widely available data across domains \cite{feng2024layoutgpt, zha2023sei}.

As shown in Fig.~\ref{fig:generator}, common generative models include Variational Autoencoders (VAEs), Generative Adversarial Networks (GANs) \cite{goodfellow2014generative}, and Diffusion Models. Diffusion models, foundational in generative modeling, demonstrate efficacy across diverse modalities including images, audio, and video.

Specifically, diffusion probabilistic models like Denoising Diffusion Probabilistic Models (DDPMs) \cite{ho2020denoising} and Stochastic Differential Equation (SDE)-based models \cite{song2020score} employ a dual-phase optimization approach. They incorporate a diffusion phase introducing noise and a denoising phase reverting this noise to generate data aligned with the training distribution. DDPMs operate in fixed, discrete steps, iterating noise addition and reversal, while SDE-based models offer continuous flexibility in handling the diffusion process, enabling sample generation at arbitrary points.

Specifically, diffusion probabilistic models like Denoising Diffusion Probabilistic Models (DDPMs) \cite{ho2020denoising} and Stochastic Differential Equation (SDE)-based models \cite{song2020score} employ a dual-phase optimization approach. They incorporate a diffusion phase introducing noise and a denoising phase reverting this noise to generate data aligned with the training distribution. DDPMs operate in fixed, discrete steps, iterating noise addition and reversal, while SDE-based models offer continuous flexibility in handling the diffusion process, enabling sample generation at arbitrary points. In industrial applications where data generation is required on demand, these models prove especially valuable \cite{ren2023dlformer}. By generating synthetic data that mirrors real-world conditions, DDPMs and SDE-based models provide the ability to quickly respond to evolving industrial needs, whether in simulation, testing, or adaptive system training.

\begin{figure}[t]
    \centering
    \includegraphics[width=0.9\linewidth]{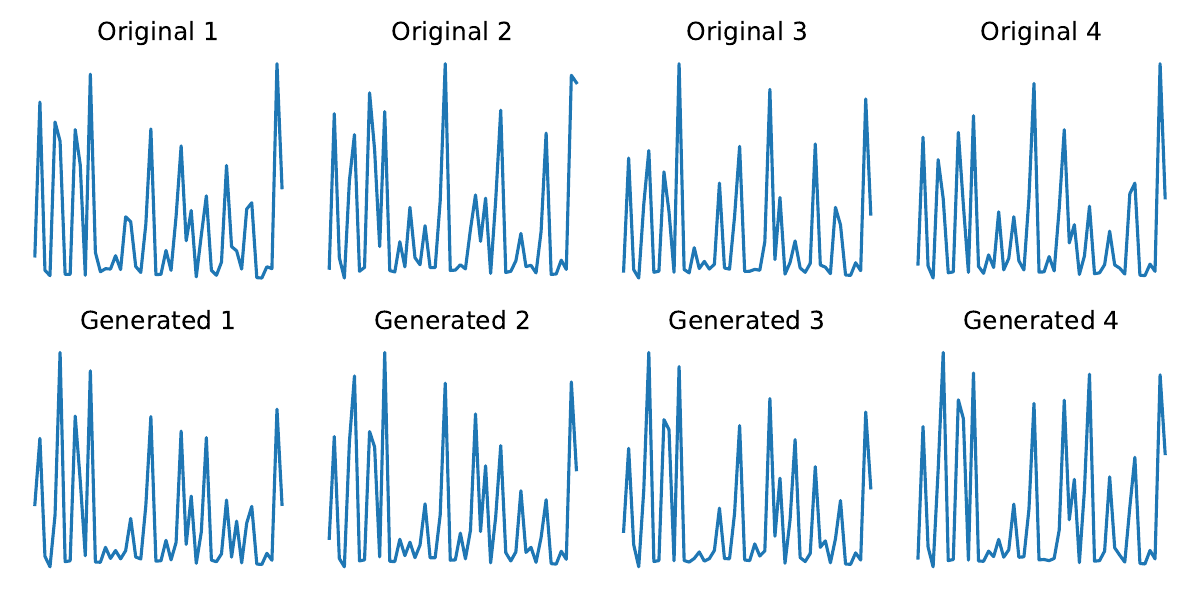}
    \caption{Comparison of raw time series samples and generative samples using DDPM \cite{ho2020denoising}.}
    \label{fig:raw_and_generative}
\end{figure}

Furthermore, we include a comparison between raw and generated data, as illustrated in Fig.~\ref{fig:raw_and_generative}, highlighting the ability of DDPM to produce samples closely resembling the original time series data.

With the increasing demand for AIGC \cite{10398474}, the deployment of AIGC, such as ChatGPT and Dall-E, has garnered widespread attention in mobile edge networks. These networks can provide personalized and customized AIGC services in real time while ensuring user privacy. However, to achieve these capabilities, it is essential to thoroughly explore the background of generative models, the lifecycle of AIGC services, and the cloud-edge-mobile collaborative required to support these services.

Beyond data generation, diffusion models find application in automating dataset construction. Tools like FABRICATOR \cite{golde2023fabricator} use them to produce labeled data for NLP tasks, while DYNOSAUR \cite{yin2023dynosaur} automates training instruction generation based on existing dataset metadata. In visual content generation, diffusion models empower the creation of intricate layouts from textual descriptions, enhancing user control without exhaustive manual inputs.

\subsubsection{Generation in IIoT}

We analyze the relationship between communication networks and large models (e.g., AIGC), and Table~\ref{tab:aigc}  clearly presents the commonalities and differences across these studies. The commonalities indicate that all the studies focus on optimizing AIGC services by utilizing communication networks and/or edge computing resources, highlighting the close relationship between communication networks and large models. The differences are analyzed across three dimensions: research focus, key technologies, and application scenarios. These differentiated analyses help readers gain a clearer understanding of the unique contributions of each study.

In conclusion, the integration of communication networks and large models is showing a trend of diversification. Various research directions and technical approaches provide valuable insights for the practical application of AIGC services. As research progresses, the collaborative development of communication networks and large models is expected to open up more possibilities for future intelligent applications.

\begin{table*}[]
\centering
\caption{Comparative analysis of research studies focusing on communication networks and AIGC.}
\label{tab:aigc}
\resizebox{0.9\linewidth}{!}{
\begin{tabular}{p{2cm}p{4cm}p{4cm}l}
\toprule
\textbf{Research}              & \textbf{Main Focus}                                                                  & \textbf{Main Technologies}                                                & \textbf{Application Scenario} \\ \midrule
EUCI \cite{10804565}                           & Inferring diffusion models for AIGC services.                                        & Model inference                                                           & Resource-constrained     \\ \midrule
AIGC-IoV \cite{10528244}                       & Bridging generative AI with vehicular networks to enhance AIGC services.             & Network integration, model deployment                                     & Urban areas                   \\ \midrule
SemCom Networks \cite{10614204}    & Inferring service requirements and enhancing network architecture for AIGC services. & Network architecture (including protocol, topology customization, etc.)   & Smart cities                  \\ \midrule
Physical Layer \cite{10623395} & Enhancing physical layer security for AI-powered communication networks.             & Physical layer security (encryption protocols, security algorithms, etc.) & Critical infrastructure       \\ \midrule
Pedestrian Detection \cite{10438990}  & Enhancing AI-driven security systems through data augmentation.                      & Data augmentation techniques                                              & Public safety                 \\ \midrule
Mobile AIGC \cite{10398474}         & Optimizing AIGC services for mobile networks.                                        & Network flexibility, efficiency, security, data management                & Mobile networks               \\ \bottomrule
\end{tabular}
}
\end{table*}

\section{IIoT: Training Ground for LMs}
\subsection{Pre-training}

\begin{figure*}[t]
    \centering
    \includegraphics[width=0.8\linewidth]{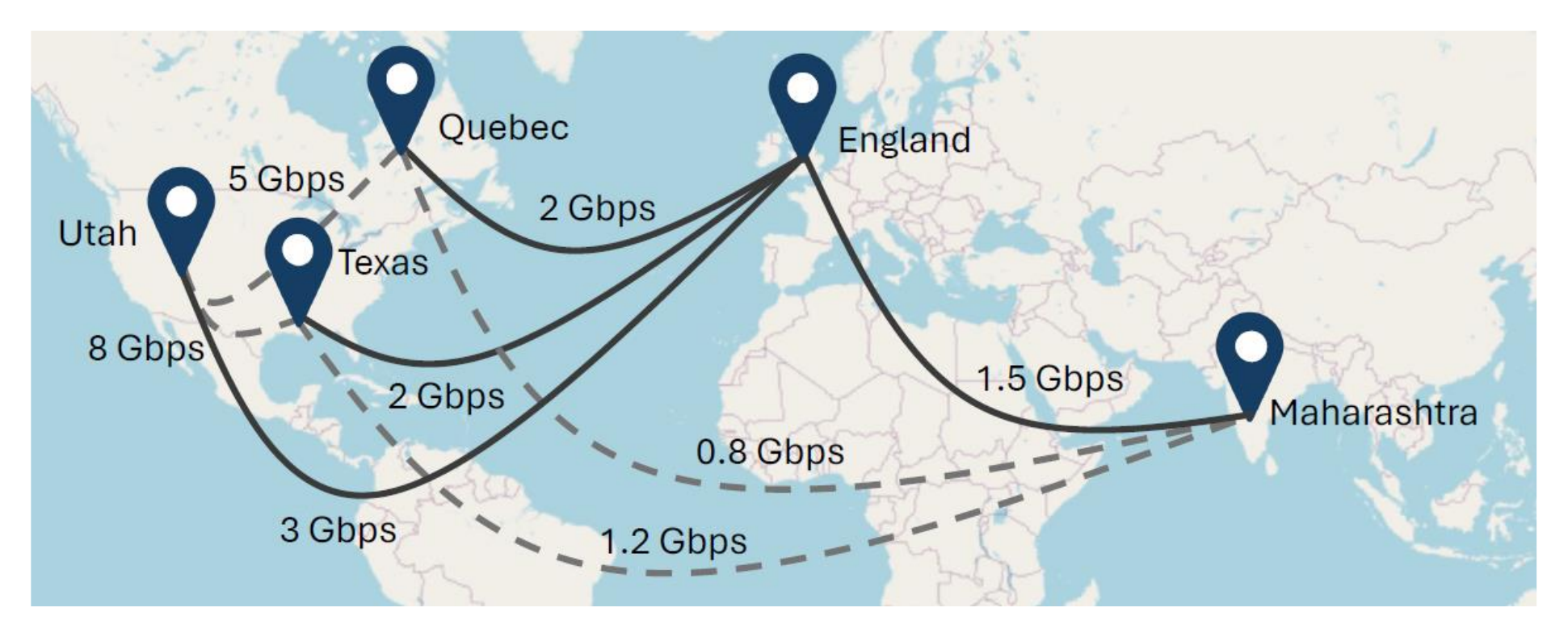}
    \caption{
    The schematic diagram of the Photon \cite{sani2024photon} system.
    }
    \label{fig:photon}
\end{figure*}

The rapid development of federated learning (FL) has significantly advanced the progress of GII \cite{pmlr-v54-mcmahan17a, 10447769, chen2022globecom, zhang2023spatial}. FL enables the training of a global model through the exchange of model parameters or gradients between a central server and edge devices, while simultaneously preserving data privacy and effectively mitigating the challenges posed by data silos. This decentralized approach has proven to be a powerful solution in scenarios where sensitive data must remain localized, offering significant advantages for real-world applications.

The integration of FL with GII shows considerable promise for future advancements \cite{MA2023102956}. When combined with IIoT, FL enhances application possibilities across various domains, including data sharing, task offloading, attack detection, localization, and privacy/security enhancement \cite{9415623,10400810}. Specifically, FL-based approaches have proven valuable in sectors such as smart healthcare, smart transportation, UAVs, smart cities, and intelligent manufacturing, where data privacy and distributed computing are critical.

IIoT devices are often widely distributed, and network quality can vary significantly, especially in cross-island federated learning scenarios. Bandwidth limitations can severely impact communication efficiency, thereby affecting training performance. Photon \cite{sani2024photon} addresses this challenge by enabling end-to-end pretraining of LLMs in low-bandwidth, globally distributed environments. Designed for collaborative training across regions and institutions, Photon outperforms centralized models by improving communication efficiency. It employs low-bandwidth optimization algorithms, such as \texttt{LocalSGD}, and adapts to varying client connection conditions with a dynamic local parallel mechanism, reducing communication overhead. Additionally, Photon compresses model updates using techniques like model quantization and sparse updates, significantly reducing the data transmitted per communication round. Fig.~\ref{fig:photon} illustrates the Photon system’s schematic.

\subsubsection{Bandwidth Limitation and Communication Efficiency}

IIoT devices are often widely distributed, and the quality of network connections can vary significantly. This is particularly evident in cross-island federated learning scenarios, where bandwidth limitations can significantly impact communication efficiency, thereby affecting training performance. To address this challenge, Photon \cite{sani2024photon}, a federated learning system, focuses on end-to-end pretraining of LLMs in low-bandwidth, globally distributed environments. Its primary goal is to enable collaborative training across regions and institutions, ultimately producing models that outperform those trained via centralized approaches. To enhance communication efficiency, Photon introduces low-bandwidth optimization algorithms such as \texttt{LocalSGD}, and combines them with an adaptive local parallel mechanism that dynamically adjusts distributed and low-bandwidth algorithms based on client connection conditions, effectively reducing communication overhead. Moreover, Photon reduces communication frequency and compresses model updates, further decreasing communication volume. It also employs techniques like model quantization and sparse updates to compress the amount of data transmitted per communication round, significantly improving the system's communication efficiency. Fig.~\ref{fig:photon} illustrates a schematic diagram of the Photon system.

The common goal of these studies is to reduce communication overhead in FL, thereby improving training efficiency, especially in environments with limited bandwidth and high latency. The research primarily focuses on wireless networks and IoT environments, which often face challenges such as bandwidth constraints and device heterogeneity, making the optimization of the FL communication process essential. To mitigate communication costs, many studies employ techniques such as client selection \cite{9656631,10181138}, bandwidth allocation \cite{9656631,10181138,10122600}, and knowledge distillation \cite{10618889, 10622906} to reduce the amount of data transferred and enhance training efficiency.

Although these studies share similar objectives and technical directions, they exhibit differences in focus and implementation. Some studies concentrate on algorithmic optimizations, such as reducing the amount of data required for model updates through knowledge distillation or optimizing bandwidth usage through joint client selection and resource allocation algorithms. In contrast, other studies emphasize communication technologies, such as clustering-based Federated Learning or utilizing multi-carrier NOMA techniques to improve bandwidth utilization and communication efficiency \cite{10753028}. Additionally, the methodological approaches differ, with some studies employing optimization theory and linear programming to solve resource allocation issues \cite{9656631}, while others use deep reinforcement learning (DRL) to dynamically adjust scheduling strategies \cite{10181138}. These differences indicate that various technical solutions are suited to different application scenarios, and selecting the appropriate approach requires considering the specific environment. We compare the relevant literature, and the results are shown in Table~\ref{tab:Bandwidth}.

\begin{table*}[!t]
\centering
\caption{Pre-training in Bandwidth and Communication Constrained IIoT.}
\label{tab:Bandwidth}
\resizebox{0.9\linewidth}{!}{
\begin{tabular}{cccccccc}
\toprule
\textbf{Method}                           & \textbf{Wireless/IoT} & \textbf{Client Selection} & \textbf{Bandwidth Allocation} & \textbf{KD}  & \textbf{DRL} & \textbf{NOMA} & \textbf{Clustering} \\ \midrule
KD-DP \cite{10618889} & $\checkmark$          &                           &                               & $\checkmark$ &              &               &                     \\ 
Fed-EC \cite{10753028} & $\checkmark$          &                           &                               &              &              &               & $\checkmark$        \\ 
FedSKD \cite{10622906} & $\checkmark$          &                           &                               & $\checkmark$ &              &               &                     \\ 
J-CSBA \cite{9656631}  & $\checkmark$          & $\checkmark$              & $\checkmark$                  &              &              &               &                     \\ 
DS-BA \cite{10181138} & $\checkmark$          & $\checkmark$              & $\checkmark$                  &              & $\checkmark$ &               &                     \\ 
MC-NOMA \cite{10122600} & $\checkmark$          &                           & $\checkmark$                  &              &              & $\checkmark$  &                     \\ \bottomrule
\end{tabular}
}
\end{table*}

\subsubsection{Latency Issues}
Network latency is one of the primary challenges in FL, as it can lead to delays in model parameter transmission and synchronization, significantly prolonging training time. To mitigate this issue, recent studies have proposed a variety of strategies, primarily focusing on asynchronous learning, adaptive optimization, and model compression. 

To provide a comprehensive overview, we summarize these recent advancements in Table~\ref{tab:latency}. The table uses the following abbreviations to describe the primary techniques employed in each study:

AA refers to asynchronous aggregation,
AS stands for adaptive strategy,
MC represents model compression,
HM denotes heartbeat mechanism,
HA indicates hierarchical aggregation,
BC refers to bias correction,
NS stands for neighbor selection, and
DRL represents deep reinforcement learning.

First, asynchronous federated learning (AFL) has been extensively studied. Its core idea is to allow clients to update model parameters asynchronously, avoiding the constraints of global synchronization and reducing sensitivity to network latency (e.g., AS-AFL \cite{10440169}, MD-FEEL \cite{10622848}). Asynchronous methods often optimize communication efficiency through local aggregation, neighbor selection, or hierarchical architectures, while employing bias correction techniques to address potential model update inconsistencies.

Second, adaptive optimization strategies dynamically adjust network parameters, such as model aggregation frequency, synchronization degree, and bandwidth allocation, to minimize communication delay and model loss (e.g., adaptive semi-asynchronous federated learning \cite{10589673}). These strategies often integrate network conditions and device resources, optimizing client selection and resource scheduling to enhance overall system efficiency.

Additionally, model compression have emerged as an essential approach to addressing network latency. Techniques such as quantization and sparsification reduce the communication overhead of model updates, significantly lowering bandwidth requirements while maintaining model convergence (e.g., QuAsyncFL \cite{10168200}). Some studies further combine these methods with hierarchical aggregation frameworks to optimize data flow and reduce cross-layer communication costs \cite{10622848}.

\begin{table}[!t]
\centering
\caption{Comparison of Approaches to Solving Network Latency.}
\label{tab:latency}
\resizebox{1\linewidth}{!}{
\begin{tabular}{lcccccccc}
\toprule
\textbf{Method} & \textbf{AA}  & \textbf{AS}  & \textbf{MC}  & \textbf{HM}  & \textbf{HA}  & \textbf{BC}  & \textbf{NS}  & \textbf{DRL} \\ \midrule
AS-AFL \cite{10440169}          & $\checkmark$                      & $\checkmark$               &                            & $\checkmark$                 &                                   &                          &                             &              \\ 
ASAFL \cite{10589673}         & $\checkmark$                      & $\checkmark$               &                            &                              &                                   &                          &                             &              \\ 
AsyDFL \cite{10599287}          & $\checkmark$                      &                            &                            &                              &                                   & $\checkmark$             & $\checkmark$                &              \\ 
MD-FEEL \cite{10622848}         & $\checkmark$                      &                            &                            &                              & $\checkmark$                      &                          &                             &              \\ 
QuAsyncFL \cite{10168200}       & $\checkmark$                      &                            & $\checkmark$               &                              &                                   &                          &                             &              \\ 
J-DLAL \cite{10041216}          & $\checkmark$                      & $\checkmark$               & $\checkmark$               &                              &                                   &                          &                             &              \\ 
Semi-AFL \cite{10193846}        & $\checkmark$                      & $\checkmark$               &                            &                              &                                   &                          &                             & $\checkmark$ \\ \bottomrule
\end{tabular}
}
\end{table}

\subsubsection{Scalability}
As the number of IIoT devices continues to grow, FL systems must be highly scalable to efficiently address the challenges posed by large-scale device participation in training. Recent studies have proposed various solutions, focusing primarily on hierarchical architectures, decentralized designs, clustering methods, and reputation mechanisms.

First, Hierarchical Federated Learning (HFL) has emerged as a key approach to enhancing scalability \cite{10167449,10439630,10091843,10046398}. By introducing intermediate aggregation nodes at different levels (e.g., cloud, edge, and device layers), HFL effectively reduces communication overhead and latency while improving model training efficiency. For instance, HFL methods are widely applied in edge computing and vehicular networks \cite{10046398}, where local aggregation at edge servers or intermediate nodes minimizes direct communication with cloud servers. These methods also incorporate clustering based on social context or device characteristics to enhance model personalization and accuracy.

Second, Decentralized Federated Learning (DFL) offers an architecture that eliminates reliance on a central server, significantly improving system robustness and fault tolerance \cite{10216376,10439630,shi2023improving}. By leveraging distributed topologies, DFL facilitates efficient model transmission and asynchronous updates among nodes, mitigating the risks associated with central server failures. Furthermore, optimization algorithms have been proposed to enhance model consistency in DFL, improving generalization performance in heterogeneous data and complex network environments.

Additionally, clustering-based participant selection methods have shown potential in improving FL efficiency. By grouping devices based on resource availability, data distribution, or social context information, clustering optimizes device allocation and model training processes. In Internet of Medical Things (IoMT) applications, clustering is used to select the most suitable participants, thereby enhancing system performance and reducing resource consumption \cite{10091843}.

Lastly, reputation mechanisms are widely employed to enhance system robustness \cite{10046398}. By dynamically evaluating participant reliability and designing robust aggregation algorithms, these mechanisms effectively defend against malicious devices or intermediate nodes. For example, hierarchical architectures incorporating reputation updates dynamically adjust model aggregation weights, further improving model security and training accuracy.

\textbf{Summary:}  
To address scalability challenges posed by large-scale device participation, current research focuses on the following strategies:
\begin{itemize}
    \item \textbf{HFL:} Reducing communication overhead and improving training efficiency through intermediate aggregation nodes.
    \item \textbf{DFL:} Eliminating reliance on central servers to enhance system robustness.
    \item \textbf{Clustering:} Optimizing participant allocation to improve model personalization and efficiency.
    \item \textbf{Reputation:} Dynamically evaluating participant reliability to enhance system security.
\end{itemize}

We present a comparative summary of the related studies in Table~\ref{tab:scalability}.

\begin{table*}[!t]
\centering
\caption{Comparison of approaches to solving scalability.}
\label{tab:scalability}
\resizebox{0.8\linewidth}{!}{
\begin{tabular}{lcccc}
\toprule
\textbf{Research Method}                        & \textbf{Hierarchical FL} & \textbf{Decentralized FL} & \textbf{Clustering} & \textbf{Reputation} \\ \hline
DAHFL \cite{10167763}                                           & $\checkmark$             &                           & $\checkmark$        &                     \\
FedLEO \cite{10216376}                                          &                          & $\checkmark$              &                     &                     \\
FedUC \cite{10439630}                                           & $\checkmark$             & $\checkmark$              & $\checkmark$        &                     \\
SC-HFL \cite{10091843}            & $\checkmark$             &                           & $\checkmark$        &                     \\
DFedSAM \cite{shi2023improving} &                          & $\checkmark$              &                     &                     \\
RoHFL \cite{10046398}                                           & $\checkmark$             &                           &                     & $\checkmark$        \\ \hline
\end{tabular}
}
\end{table*}

\subsubsection{Security Concerns}
Security risks in IIoT environments are relatively high, as model parameters may be susceptible to attacks during transmission, leading to data leakage or model tampering. To address this, techniques like encryption and differential privacy are employed to protect model parameters \cite{10637763,9409743}. Additionally, blockchain technology is proposed to enhance data security and trustworthiness, ensuring transparent, tamper-proof transmission of model parameters and thereby improving the overall security of federated learning systems.
Ddeep learning methods can be applied to develop enhanced security solutions, further improving the ability to defend against potential attacks and boosting system robustness \cite{9072101}.

\subsection{LMs as the Tuning Predictor}\label{sec:LMs as a Tuning Predictor}

During the pre-trained phase, LMs typically acquire the general capability to handle a variety of tasks. However, to effectively deploy these pre-trained models in industrial environments, it is essential to fine-tune them according to specific task requirements, the evolving data distribution, and diverse user preferences. Fine-tuning involves further training the pre-trained model on a specific dataset to better adapt it to the demands of a particular task or domain. This process not only facilitates the infusion of domain-specific knowledge but also aligns the model with specific task instructions.

\subsubsection{The Goal of Fine-tuning}\label{sec:Alignment Fine Tuning}
Fine-tuning LLMs serves multiple objectives, including adapting models to specific tasks, correcting errors, and ensuring alignment with human values. These objectives can be categorized into three key areas: Instruction Tuning, Model Refinement, and Model Alignment, each addressing distinct challenges and advancements in fine-tuning methodologies.

\textbf{Instruction Tuning.} \label{sec:Instruction Tuning}
Instruction Tuning (IT) is a crucial technique for adapting pre-trained LLMs to specific tasks and user preferences in various domains, ensuring alignment with nuanced requirements.

Consider a LM $h(x)$ designed to process input data $x$, which typically comprises commands or queries in natural language. IT is a targeted training approach focused on enhancing the model's precision and efficiency in handling specific directives. The objective of IT involves the recalibration of $h$ by fine-tuning its parameters utilizing a carefully selected training set $\mathcal{I} = {(x_i, \hat{y}_i)}_{i=1}^{N}$. In this set, $\hat{y}_i$ denotes the anticipated response for the input $x_i$, emphasizing improvements in specific tasks or functionalities. The redefined model, noted as $h'$, operates under the following rule:

\begin{equation}
    h'(x_0) = \hat{y}_0, \quad \forall (x_0, \hat{y}_0) \in \mathcal{I}.
\end{equation}

In the process of IT, user requests (Instructions) and machine responses (Answers) are synthesized into a chat-style task. This typically involves inserting tags that represent the roles of the user and machine, such as ``[USER]" and ``[BOT]", and adding special markers at the beginning and end of the dialogue, such as $<\text{bos}\_\text{token}>$ and $<\text{eos}\_\text{token}>$. The resulting concatenated text appears as a continuous dialogue stream, allowing the model to perform autoregressive predictions on this text structure.

\begin{figure}[t]
    \centering
    \includegraphics[width=1\linewidth]{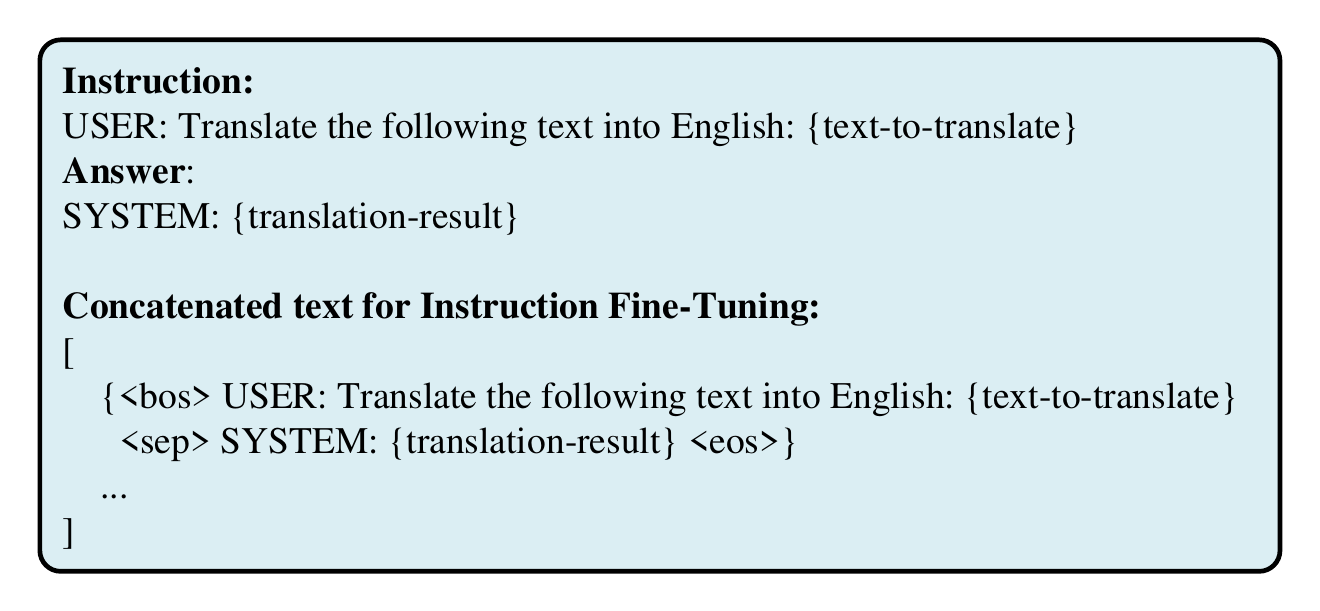}
    \caption{
    An example of instruction following data.
    }
    \label{fig:instruction_following_data}
\end{figure}

The primary distinction between the IT process and general pre-trained lies in the method of loss calculation. Although cross-entropy remains the loss function of choice, during the IT process, the loss is calculated solely based on the machine's response (i.e., the text following ``[BOT]:"). The part corresponding to the user request is excluded from the loss calculation by setting the ignore\_index parameter. Fig.~\ref{fig:instruction_following_data} shows a training sample of IT.

Faysse \textit{et al.} \cite{faysse2023revisiting} provide a detailed exposition of the various evaluation metrics employed during directive fine-tuning for adaptation to downstream tasks, offering practitioners actionable insights for the practical deployment of directive fine-tuned models.
By organizing two different types of dialogue task flows, InstrDialog and InstrDialog++, Zhang \textit{et al.} \cite{zhang2023citb} investigate the problem of Continual Instruction Tuning (CIT). The study demonstrates that sequentially fine-tuning instruction-tuned models yields performance competitive with existing CL methods. Furthermore, rich natural language instructions facilitate knowledge transfer and mitigate forgetting, an advantage that current CL methods scarcely leverage.

\textbf{Model Refinement.} \label{sec:Model Refinement}
Like humans, LLMs are prone to errors, such as producing inaccurate translations or retaining outdated information. Ensuring that LLMs remain accurate and up-to-date typically involves direct fine-tuning of the model. While effective, this method can be time-consuming and risk degrading performance on tasks the model has previously learned. To overcome these challenges, the concept of Model Refinement (MR) \cite{sinitsin2020editable} has been developed as a technique to correct specific errors without disrupting the primary functionalities of the model. This approach, initially guided by a ``reliability-locality-efficiency" principle, implements an efficient editing process using gradient descent. Subsequent research \cite{de2021editing, mitchell2021fast} have expanded on this foundation, applying gradient decomposition techniques to larger models such as GPT-J-6B and T5-XXL. These methods focus on updating a subset of model parameters to modify the labels of specific inputs and integrate editing through retrieval mechanisms.

Let $h(x)$ be a model that processes input data $x$, such as natural language queries. The model is subjected to an editing set $\mathcal{E} = {(x_e, y_e, \hat{y}_e)}_{e=1}^{N}$, wherein $\hat{y}_e$ is the correct label for $x_e$, yet the model erroneously predicts $y_e$. The purpose of MR is to modify $h$ into $h'$, improving the model's accuracy on the editing set $\mathcal{E}$. The goal is for $h'$ to accurately predict the correct labels for inputs in $\mathcal{E}$ while preserving its original performance on inputs not included in $\mathcal{E}$. The refined model $h'$ functions according to the following definition:

\begin{equation}
    h'(x_0) = 
    \begin{cases}
        \hat{y}_0 & \text{if } (x_0, \hat{y}_0) \in \mathcal{E},\\
        h(x_0) & \text{otherwise}.
    \end{cases}
\end{equation}

Although these innovations lay a solid foundation for research in Continuous Model Refinement (CMR), the field of continual refinement for LLMs remains largely unexplored. A study has highlighted a potential issue: the optimal location for storing factual data might not coincide with the most effective place for editing it \cite{hase2024does}. This discrepancy challenges the traditional "locate and edit" paradigm and could pose significant concerns in the implementation of CMR. Moreover, questions about whether this approach is suitable for LLMs and whether more memory- and computationally efficient methods can be developed still demand further exploration and resolution.

\textbf{Model Alignment.} \label{sec:Model Alignmen}
Model Alignment (MA) is a fundamental concept in the deployment of AI systems, aimed at ensuring that the actions and outputs of AI models are in harmony with human values, ethics, and preferences. This process involves the use of mathematical models, algorithmic adjustments, and iterative feedback to refine AI behaviours, enabling the model to adapt continuously to a stream of incoming data and remain up-to-date. 

Consider $h(x)$ as a model tailored for decision-making scenarios, processing inputs $x$. An alignment dataset $\mathcal{A} = {(x_a, y_a, \hat{y}_a)}_{a=1}^{M}$ is utilized, in which $y_a$ is the model’s initial decision for the input $x_a$, and $\hat{y}_a$ is the corrected decision conforming to established ethical standards or desired outcomes. MA focuses on transforming $h$ into $h'$, ensuring that $h'(x_a)$ yields $\hat{y}_a$ for any $x_a$ from the alignment dataset, thus realigning the model’s decisions according to the specified criteria. The adjusted model $h'$ operates under the following rule:

\begin{equation}
    h'(x_0) = \hat{y}_0, \quad \forall (x_0, \hat{y}_0) \in \mathcal{A}.
\end{equation}

Current research efforts focus on developing methods to minimize the challenges associated with model alignment \cite{rafailov2024direct, puthumanaillam2024moral}. These methods implement specific optimization strategies that enhance the adaptability of LLMs, ensuring their outputs remain ethically and factually relevant within dynamic societal and information landscapes.

Furthermore, frameworks for Continuous Model Alignment (CMA), based on Reinforcement Learning (RL) and Supervised Learning (SL), have been proposed. The RL-based CMA framework aims to reduce alignment costs and prevent forgetting by employing more efficient reinforcement learning techniques \cite{ouyang2022training}. The SL-based CMA framework integrates AI systems through direct supervision on dynamic datasets, focusing on maintaining long-term alignment with human values \cite{zhangcppo}. These frameworks demonstrate how continuous learning and model adjustments can be applied in practical industrial settings, thereby enhancing the operational efficiency and decision-making quality of AI systems.

\subsubsection{Parameter Efficient Fine-tuning}\label{sec:Efficient Tuning}
\begin{figure}[t]
    \centering
    \includegraphics[width=0.9\linewidth]{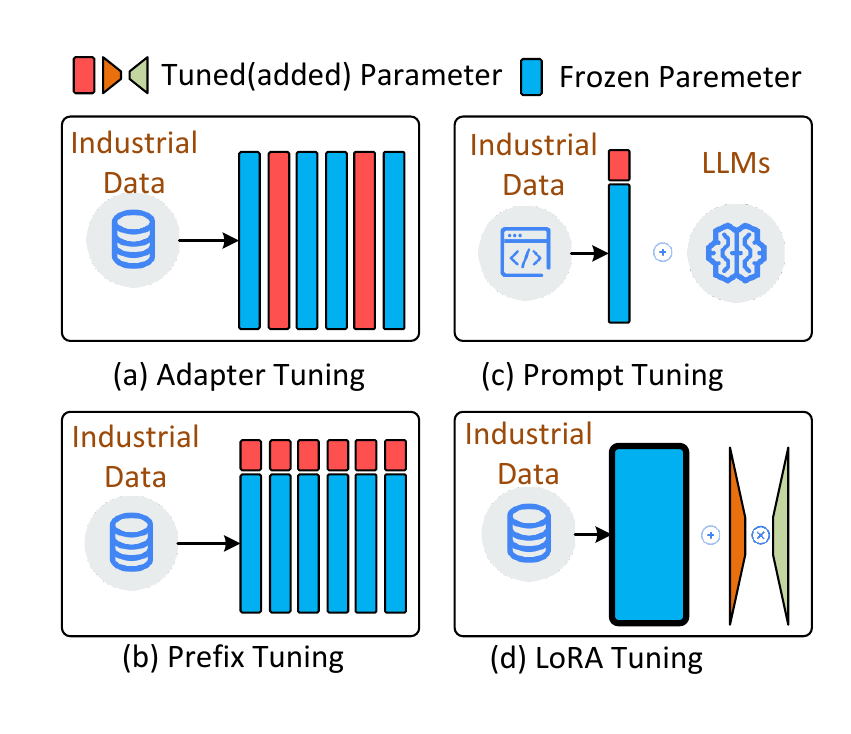}
    \caption{
    Illustrations of parameter-efficient fine-Tuning.
    }
    \label{fig:peft}
\end{figure}

Efficient fine-tuning aims to improve the efficiency of fine-tuning processes for LMs.
Illustrated in Fig.~\ref{fig:peft}, efficient fine-tuning methods are categorized as Density-Efficient Fine-Tuning, Memory-Efficient Fine-Tuning, and Parameter-Efficient Fine-Tuning (PEFT).

\begin{figure}[t]
    \centering
    \includegraphics[width=0.8\linewidth]{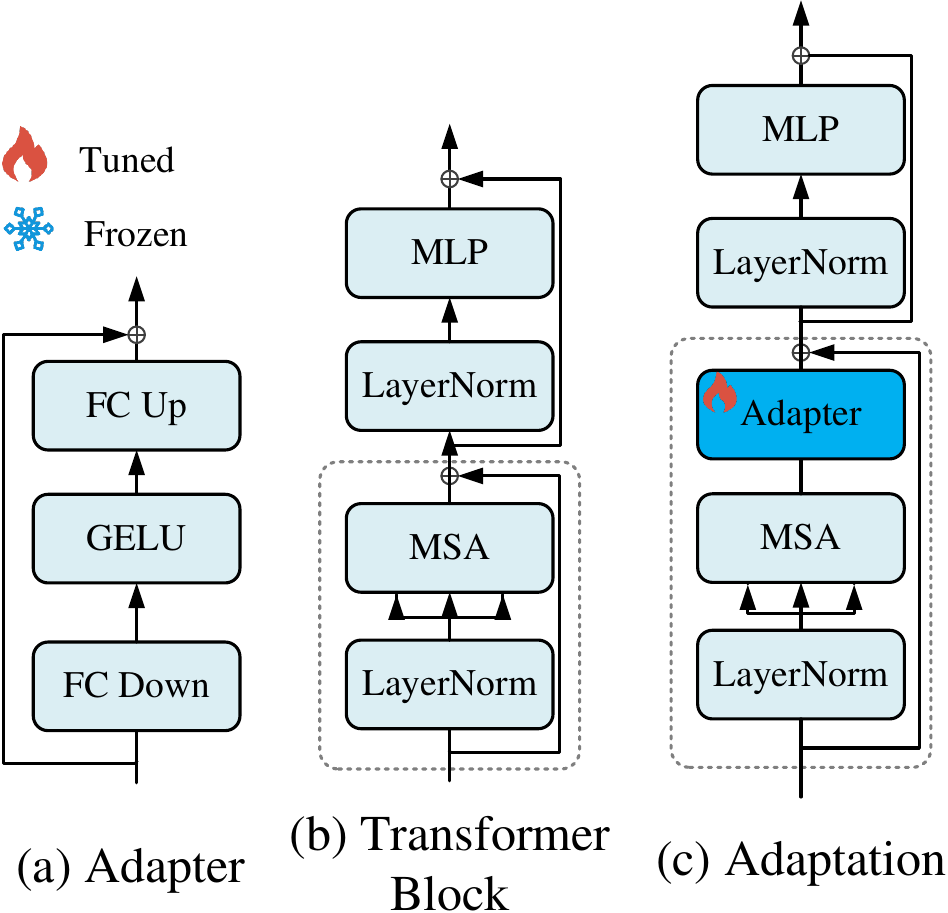}
    \caption{
    Adding adapters to Transformers. During training, only newly added Adapters are updated while all the other layers are frozen.
    }
    \label{fig:adapter}
\end{figure}

\textbf{Adapter Tuning.} 
Adapter Tuning entails adding extra small neural network layers to specific Transformer layers within the pre-trained model.
Houlsby \textit{et al.} \cite{houlsby2019parameter} design a bottleneck module that initially reduces the dimensionality of the original features, applies a nonlinear transformation, and then restores them to their original dimension.
The bottleneck module is integrated into each Transformer layer, usually inserted sequentially after the Transformer layer's two core components, namely the attention and feed-forward layers, as shown in Fig.~\ref{fig:adapter}.
During fine-tuning, only the weights of these adapters are updated, leaving the original model parameters unchanged.
This effectively directs the model's learning capacity towards the adapters to facilitate adaptation to new tasks.

\textbf{Prefix Tuning.} 
Prefix Tuning involves adding a series of learnable continuous vectors (referred to as prefixes) to the front of each Transformer layer of the model \cite{li2021prefix}. 
These prefixes serve as task-specific signals that guide the model in generating task-oriented responses while keeping most parameters unchanged.
Jia \textit{et al.} \cite{jia2022visual} introduce Visual Prompt Tuning as an efficient alternative to fully fine-tuning large Transformer models in vision tasks.
VPT introduces a small number (less than 1\% of the model parameters) of trainable parameters in the input space to adapt to various downstream recognition tasks, achieving significant performance improvements.

\textbf{Prompt Tuning.} 
Prompt tuning primarily adds trainable prompt vectors at the input layer of the model.
Specific input texts (i.e., prompts) are chosen or generated using discrete prompt methods to enrich the input text. The augmented input is used to produce model outputs beneficial for downstream tasks \cite{jiang2020can}.
Recent research has also introduced soft prompt methods, which guide frozen LMs in performing specific downstream tasks by directly incorporating prefix prompts into the input \cite{lester2021power}.
Soft prompts are learned via backpropagation and can be fine-tuned with labelled examples to include signals from end-to-end learning. 
Wang \textit{et al.} \cite{wang2022learning} applied the concept of prompt tuning to CL by creating a prompt pool and fetching prompts relevant to the current task to prevent catastrophic forgetting.

\textbf{LoRA.} 
LoRA \cite{hu2021lora} accomplishes fine-tuning by altering the weights of the pre-trained model through low-rank decomposition.
This technique entails substituting the weight matrices of the model's linear layers with the product of two low-rank matrices, thereby decreasing the number of parameters requiring updates during fine-tuning.
The original weight matrices remain unchanged, with only the introduced low-rank matrices being adjusted during training.

\subsubsection{Other Fine-tuning}
To address the growing computational and memory challenges of fine-tuning LLMs, recent research has proposed innovative approaches.

\textbf{Density Efficient Fine-tuning.} 
With increasing model sizes, traditional fine-tuning of all parameters becomes increasingly challenging.
PEFT methods have gained popularity for effectively adapting pre-trained LMs.
Recent research has identified the presence of activation sparsity within the Multilayer Perceptron (MLP) blocks of transformers.
Lower activation density enables efficient model inference on hardware optimized for sparsity.
Building on this insight, a new method called DEFT \cite{runwal2024peft} has been proposed to reduce loss by promoting higher activation sparsity (or lower activation density) in pre-trained models.

\textbf{Memory Efficient Fine-tuning.}
Traditional fine-tuning methods rely on backpropagation algorithms. However, as the model size increases, they demand exponentially more memory.
Zeroth-order methods theoretically estimate gradients with only two forward passes but are typically very slow when optimizing LMs.
MeZO \cite{malladi2024fine} enhances the traditional Zeroth-order Stochastic Gradient Descent method by adapting it to in-place operations, maintaining the memory usage for fine-tuning LMs at the same level as that during inference.

\subsection{LMs as the Non-Tuning Predictor}\label{sec:LMs as a Non-Tuning Predictor}
Non-tuning predictors for closed-source models involve structured preprocessing of industrial data to ensure compatibility with the input requirements of LLMs. This process is critical in environments where model parameters cannot be tuned. The process is divided into two main stages:

$\bullet$ \textbf{Data Preprocessing.}
Raw industrial data undergoes preprocessing to be suitably formatted for LLMs. This preprocessing typically includes operations such as:
\begin{equation}
    \begin{split}
        \hat{X}_{\text{in}} &= \text{Template}(X, P), \\
        \hat{X}_{\text{in}} &= \text{Tokenizer}(X),
    \end{split}
\end{equation}
where $X$ represents the raw data, and $P$ is the instructional prompt specifying the analysis task's requirements. The $\text{Template(·)}$ structures the data according to predefined formats while the $\text{Tokenizer(·)}$ converts the data into a sequence of textual tokens suitable for LLM processing.

$\bullet$ \textbf{Model Interaction and Response Generation. }
The preprocessed data, $\hat{X}_{\text{in}}$, is input into the LLM, denoted as $f_{\text{LLM}}$. The model processes these inputs to generate outputs, which are then parsed to extract predictive labels:
\begin{equation}
\hat{Y} = \text{Parse}(f_{\text{LLM}}(\hat{X}_{\text{in}})),
\end{equation}
where $\hat{Y}$ denotes the labels predicted by the LLM based on the processed inputs.

To further enhance understanding, we review prompt-based and tokenizer-based methods that utilize LLMs effectively in non-tuning scenarios.

\begin{figure}[t]
    \centering
    \includegraphics[width=0.9\linewidth]{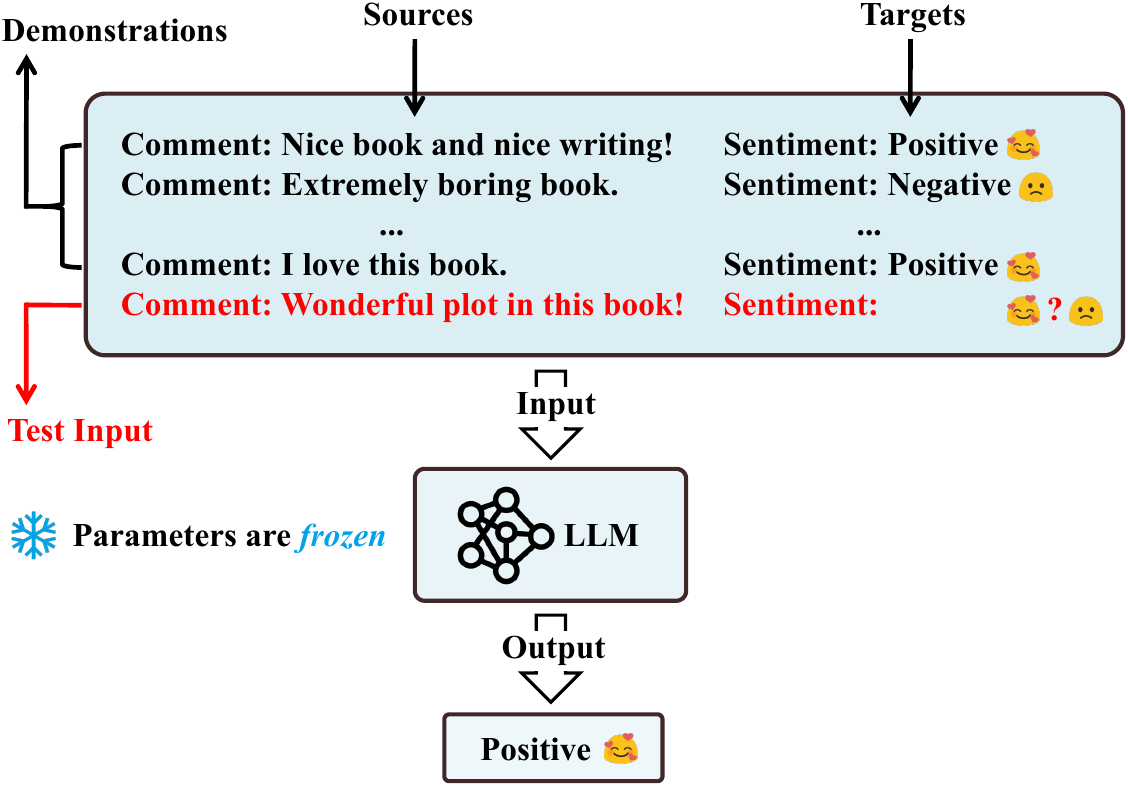}
    \caption{
    In-context Learning for classification tasks. 
    }
    \label{fig:non-tuning}
\end{figure}

\subsubsection{Prompt-based Methods.} A notable method translates natural language instructions into executable robot actions \cite{10235949}. This approach proposes customizable input prompts that seamlessly integrate into robotic execution systems or visual recognition programs, facilitating multi-step task plans while mitigating token limitation constraints of ChatGPT. In autonomous driving, LimSim++ \cite{fu2024limsim++} provides diverse types and modalities of prompt inputs, extracting road network and vehicle surroundings information at each decision frame, which is then conveyed in natural language to the driving agent (e.g., GPT-4).

\subsubsection{Tokenizer-based Methods.} The introduction of LiDAR-LLM \cite{yang2023lidar} leverages LLMs' reasoning capabilities to comprehend outdoor 3D scenes fully. Addressing the scarcity of paired 3D LiDAR and text data, a three-stage training strategy generates datasets that align the 3D modality with the language embedding space of LLMs. Initially, LLM tokenizers were not designed to handle numerical data or recognize associations in time series effectively. This challenge was addressed through lightweight embedding layers and refined prompt engineering \cite{spathis2023first}. LLMTime \cite{gruver2024large} utilizes innovative tokenization techniques, transforming tokens into flexible, continuous-valued representations, enabling unadjusted LLMs to achieve or surpass traditional performance in zero-shot prediction tasks.

Transitioning to in-context learning, we explore its relevance and applications in non-tuning LLM scenarios.

\subsubsection{In-Context Learning (ICL).} Introduced in GPT-3 \cite{brown2020language}, ICL, initially popularized by GPT-3, enables LLMs to make accurate predictions using minimal demonstrations, bypassing the need for gradient updates. This update-free feature not only lowers adaptation costs but also broadens the model's applicability across various tasks. The efficacy of ICL hinges on both the pre-training of the model on extensive datasets and the careful selection of demonstrations for task-specific predictions. Studies have identified several critical factors influencing the performance of ICL, including the label space, input text distribution, and the format of demonstrations \cite{min2022rethinking}. Moreover, structural similarity, diversity, and complexity within demonstrations are also significant \cite{an2023context}. Recent research by Coda-Forno \textit{et al.} \cite{coda2024meta} has shown that the capabilities of LLMs in ICL can be recursively enhanced through the ICL process itself.
Fig.~\ref{fig:non-tuning} shows an example of in-context learning. Several demonstrations containing both sources (comments for a book) and target classes (sentiments for the book) are fed into the LLM with the test input/query that only includes comment. The LLM can learn from the demonstrations and predicts the result without updating its parameters.

\section{IIoT: Connectivity Nexus for Emergence}
With the ongoing development of the IIoT, its provision of stable and reliable connectivity has created new opportunities for resource-constrained devices to support multitasking \cite{shao2024ai}. Conventional IoT devices, due to their limited computational resources, are typically restricted to handling a single task at a time. For example, drones often face challenges in concurrently performing tasks such as vehicle tracking, traffic accident detection, and area intrusion monitoring. Recent advancements in modular deep learning, combined with the robust connectivity enabled by IIoT, have facilitated the execution of multiple tasks on a single resource-limited device, thus addressing longstanding limitations in multitask processing within such environments.
\begin{figure*}[!th]
    \centering
    \includegraphics[width=0.7\linewidth]{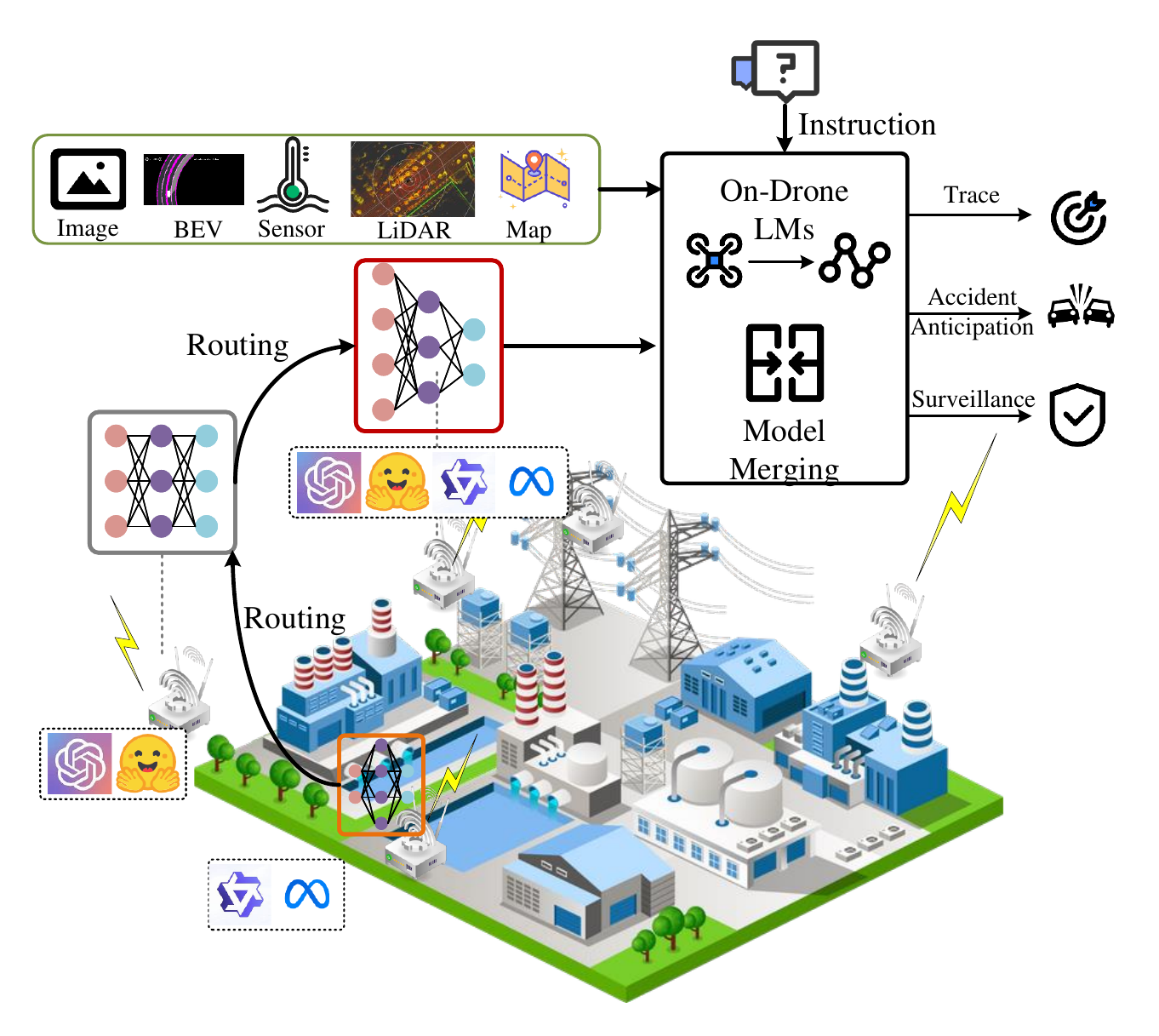}
    \caption{IIoT-enabled framework for modular routing and model merging on drones, enabling efficient multitasking.}
    \label{fig:lisa}
\end{figure*}

\subsection{Modular LMs}
In the IIoT context, the volume and variety of data generated by devices are enormous. Modular LMs enable specific modules to learn and optimize the processing of certain types of data. For example, one module may specialize in processing audio data from production lines, while another focuses on analyzing visual data. This specialization not only enhances data processing efficiency but also facilitates the reuse of learned knowledge across modules, reducing resource consumption and improving overall system performance.

\subsubsection{Modularization Facilitates Positive Transfer}
By modularizing LMs, specific tasks or domain knowledge can be encapsulated within independent modules. This encapsulation not only preserves task-specific knowledge but also promotes positive transfer—where learning in one context enhances performance in another related but different context. This strategy effectively isolates and utilizes domain-specific insights, minimizing interference caused by irrelevant data and reducing the risk of negative transfer, where performance in one task could be adversely affected by learning from another.

\subsubsection{Achieving System Generalization}
The modular architecture also supports system generalization by decoupling computation from routing. This allows the model to dynamically select and combine modules based on the incoming data. For instance, in the given scenarios, an LM on a drone might connect to different data providers or model providers through IIoT to enhance its capability on a single task. Moreover, this architecture enables the integration of functionalities to execute multiple tasks simultaneously using the same system infrastructure, such as a drone adapting its modules based on various sensor inputs and immediate demands, from monitoring crop health to anticipating industrial accidents.

Fig.~\ref{fig:lisa} demonstrates how the integration of modular deep learning with the stable connectivity provided by the IIoT enables resource-constrained devices, such as drones, to execute multiple complex tasks simultaneously. Traditionally, IoT devices are limited by computational resources and can only handle single tasks, such as vehicle tracking or accident detection. However, leveraging IIoT's reliable connectivity, data collected from diverse sensors (e.g., images, LiDAR, and maps) can be dynamically routed to task-specific expert models using modular routing mechanisms. Furthermore, model merging techniques consolidate the functionalities of multiple models into a single framework, reducing computational overhead while enhancing multitasking capabilities. Supported by IIoT's distributed infrastructure, this framework facilitates efficient real-time execution of tasks, including surveillance, accident anticipation, and trajectory analysis, providing a scalable and intelligent foundation for industrial applications.

Research in this area primarily focuses on two key techniques: \textbf{model routing} and \textbf{model merging}.

\subsection{Model Routing in IIoT}

Model routing refers to dynamically selecting the appropriate module or expert model based on the characteristics of the input data. This mechanism is analogous to the thalamus in the human brain, which routes sensory inputs to specific cortical regions. Routing offers the advantage of task-specific module selection, improving computational efficiency and reducing costs. Key advancements in this area include:

\begin{itemize}
    \item \textbf{Spatiotemporal Routing:} Temporal Netgrid models \cite{8688478} optimize routing paths in satellite networks by predicting data transmission delays based on spatiotemporal characteristics.
    \item To select the most suitable LLM for a new task from a pool of candidates, a router model has been proposed \cite{shnitzer2023large}. This data-driven approach leverages information from benchmark datasets to effectively guide the selection of the optimal LLM.
    \item \textbf{Embedding Space-Based Routing:} Utilizing nearest-neighbor algorithms to predict model performance based on the distance between task samples in embedding space, enabling dynamic model selection \cite{wang2022learning}.
    \item \textbf{Context-Aware Routing:} The Tryage framework employs a routing mechanism that selects the best expert model based on user prompts and constraints such as model size and latency, achieving superior performance on heterogeneous datasets \cite{hari2023tryage}.
    \item \textbf{Hard and Soft Routing:} Modular deep learning \cite{pfeiffer2023modular} leverages hard routing (binary selection) and soft routing (probabilistic selection) to optimize computational efficiency.
\end{itemize}

\subsection{Model Merging in IIoT}

Model merging integrates the capabilities of multiple models into a single framework, reducing the computational and storage overhead required for multitasking. Key advancements in model merging include:

\begin{itemize}
    \item \textbf{TIES-Merging:} A weighted averaging strategy resolves interference among models to improve performance on tasks like text classification and question answering \cite{yadav2023ties}.
    \item \textbf{LoRM:} A closed-form merging approach optimizes parameter-efficient modules (e.g., LoRA) through iterative freezing and merging cycles \cite{salami2024closed}.
    \item \textbf{Dataless Fusion:} This method averages model weights without additional training data, enhancing performance on downstream tasks \cite{jindataless}.
    \item \textbf{Task Arithmetic:} Task vectors are linearly combined to integrate multiple task functionalities into a single model, supporting task-specific updates and targeted adjustments \cite{ilharcoediting}.
    \item \textbf{Twin-Merging:} This approach dynamically integrates expert models by learning twin vectors and routing networks, achieving superior performance on benchmark datasets \cite{lutwin}.
\end{itemize}

\textbf{Discussion.} Model routing and model merging are complementary techniques that enable multitasking on resource-constrained devices. While routing dynamically assigns tasks to optimal modules, merging consolidates redundant models to minimize computational overhead. We summarize the work on model routing and model merging discussed in this paper in Table~\ref{tab:modular}, providing a structured overview of these techniques. Future research can explore more efficient merging strategies and adaptive module integration in dynamic environments, further advancing intelligent applications.

\begin{table*}[!t]
\centering
\caption{Modular LMs in IIoT.}
\label{tab:modular}
\resizebox{1\linewidth}{!}{
\begin{tabular}{lccl}
\toprule
\textbf{Research Method}                          & \textbf{Routing} & \textbf{Merging} & \textbf{Features}                                                                            \\ \midrule
Tryage \cite{hari2023tryage}                                            & $\checkmark$     &                  & Context-aware routing, inspired by the thalamus in the human brain.                                \\ 
Router \cite{shnitzer2023large}       & $\checkmark$     &                  & Leverage information from benchmark datasets to guide the selection of LLMs. \\ 
L2P \cite{wang2022learning}       & $\checkmark$     &                  & Routes based on the distance between new task samples and existing samples in the embedding space. \\ 
TNM \cite{8688478} & $\checkmark$     &                  & Used in satellite networks, selects routing paths based on spatiotemporal characteristics.         \\ 
ModDL \cite{pfeiffer2023modular}       & $\checkmark$     &                  & Comprehensive discussion of hard and soft routing mechanisms, e.g., linear hypernetworks.           \\ 
TIES-Merging \cite{yadav2023ties}                                      &                  & $\checkmark$     & Resolves model merging conflicts, uses weighted averaging.                                         \\ 
LoRM \cite{salami2024closed}                                              &                  & $\checkmark$     & Designed for parameter-efficient merging, e.g., LoRA modules.                                      \\ 
RegMean \cite{jindataless}                         &                  & $\checkmark$     & Merges model weights using weighted averaging without requiring additional training data.          \\ 
Task Arithmetic \cite{ilharcoediting}                                   &                  & $\checkmark$     & Integrates task-specific functionalities into a single model via task vector combinations.         \\ 
Twin-Merging \cite{lutwin}                                      &                  & $\checkmark$     & Dynamically integrates expert models using learned twin vectors.                                   \\ \bottomrule
\end{tabular}
}
\end{table*}

\subsection{Edge Test Time Adaptation}
\subsubsection{Test Time Adaptation}\label{sec:Test Time Adaptation}
\begin{figure}
    \centering
    \includegraphics[width=1\linewidth]{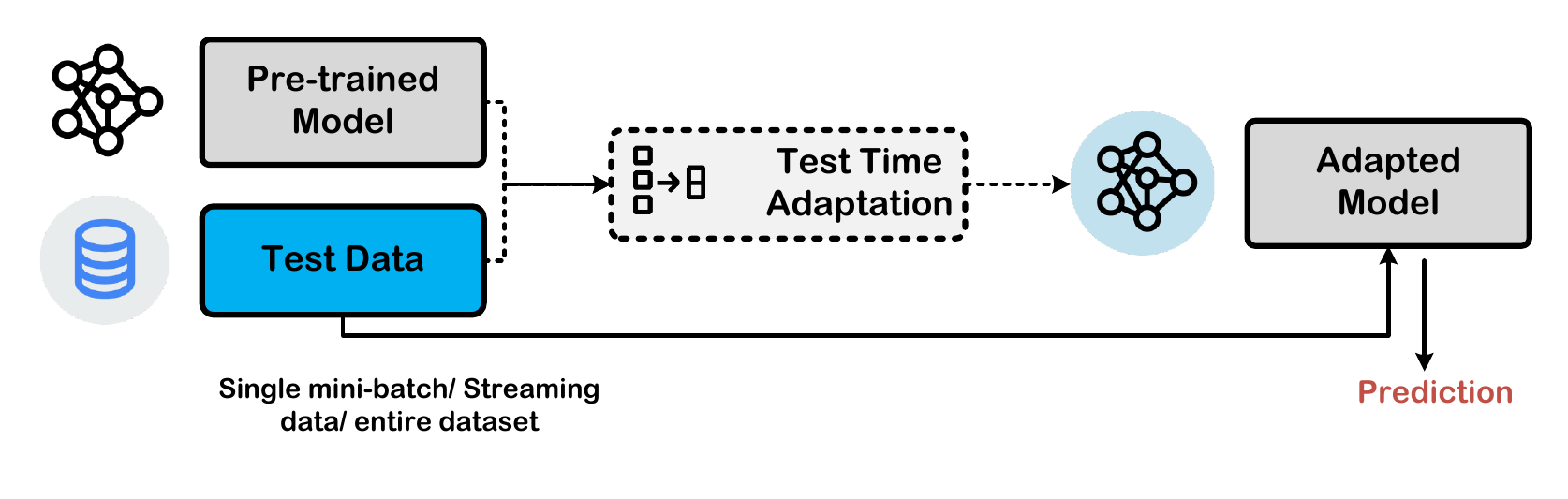}
    \caption{Test Time Adaptation.}
    \label{fig:tta}
\end{figure}

In Edge AI applications, model performance often deteriorates due to distribution shifts between training and testing data, which leads to suboptimal performance in the target domain. 
As shown in Fig.~\ref{fig:tta}, Test Time Adaptation (TTA) has emerged as an effective solution \cite{niu2024test}. TTA leverages test data to perform online adjustments of model parameters, aiming to enhance adaptability in dynamic environments. In the context of edge computing, the heterogeneous and dynamic nature of data distributions further amplifies this challenge, making TTA a critical research focus in the field.

Currently, research on TTA primarily focuses on several key aspects. First, to address the constraints of limited computational resources on edge devices, researchers aim to design efficient model update strategies. For instance, some studies propose updating only specific parameters \cite{wangtent,zhang2022memo}, such as those in normalization layers, to significantly reduce computational and memory overhead. Other approaches selectively upload valuable samples to the cloud for model updates, thereby reducing communication costs while improving update efficiency \cite{chentowards}.

However, TTA processes may encounter catastrophic forgetting, where the model loses previously learned knowledge while adapting to new distributions. 
Continual Test-Time Adaptation addresses the challenge of adapting models to continuously evolving target domains. The introduction of the Visual Domain Adapter (ViDA) \cite{liu2023vida} allows the model to handle both domain-specific and shared knowledge. The Homeostatic Knowledge Allotment strategy optimizes the integration and utilization of knowledge from different ViDA, enhancing the model's adaptability and preserving its plasticity across varying domains.
Additionally, as TTA typically operates in an unsupervised manner, prediction confidence often fails to accurately reflect true model accuracy \cite{qiu2021source,liang2020we}. To address this, some studies adopt model calibration techniques, such as consistency loss and min-max entropy regularization, to improve the alignment between predicted probabilities and actual performance.

Despite notable progress, several critical challenges remain in the field of TTA for Edge AI. Future research directions include developing lightweight TTA algorithms tailored for resource-constrained devices such as microcontrollers and sensor nodes; designing efficient TTA methods for specific applications, such as object detection, semantic segmentation, and speech recognition; and integrating TTA with federated learning frameworks to enable collaborative model updates across multiple edge devices, further improving generalization and robustness.

\subsubsection{Edge Cloud Collaboration}\label{sec:Edge Cloud Collaboration}
Edge-cloud collaboration addresses the challenges of deploying large models in resource-constrained edge environments by enabling cooperative inference and dynamic model adaptation. The Edge-Cloud Learning Model (ECLM) \cite{zhuang2023eclm} employs a modular architecture, allowing personalized edge models to be derived from cloud-based large models. This approach minimizes parameter conflicts, providing efficient adaptation and scalability for diverse industrial applications.

Resource constraints often prevent the deployment of large models locally, resulting in inference delays \cite{9866918} that hinder real-time IIoT applications such as autonomous driving, the metaverse \cite{10273380}, and digital twins \cite{10495817}. To address this, edge-cloud collaborative methods leverage cooperative inference between edge and cloud models \cite{wang2024cloud, chen2024edge}. By employing hard example mining, complex tasks are selectively offloaded to cloud-based large models for detailed processing, while simpler tasks are handled locally by lightweight edge models. This strategy ensures low latency for real-time applications while improving the accuracy of challenging tasks.

\begin{figure}[t]
    \centering
    \includegraphics[width=1\linewidth]{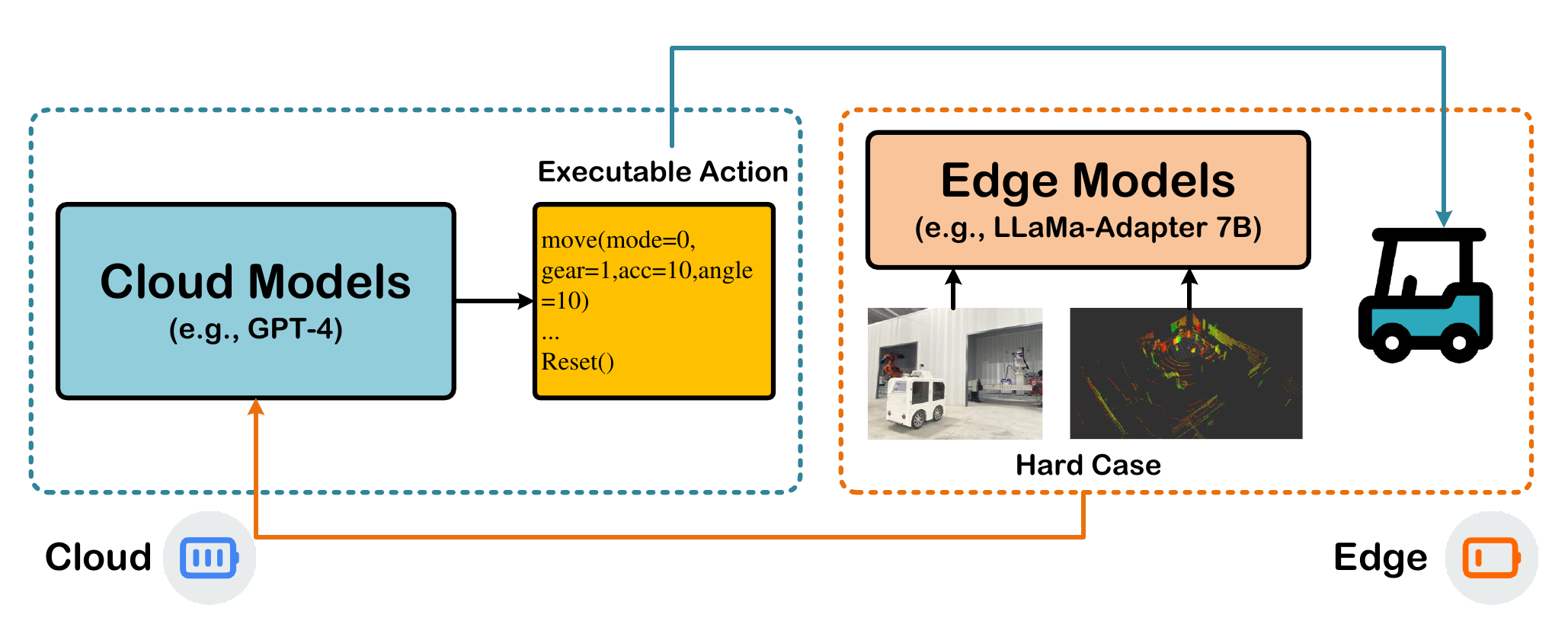}
    \caption{Edge-Cloud Collaborative Driving Planning with LMs.
    }
    \label{fig:ec}
\end{figure}

As illustrated in Fig.~\ref{fig:ec}, cloud-based large models are critical in handling complex scenarios, such as high object density or adverse visual conditions in edge-cloud collaborative systems. By offloading computationally intensive tasks to the cloud, these systems enhance overall performance, improving accuracy, reliability, and safety.

Recent frameworks emphasize integrating resources to balance responsiveness and computational capability. The Cloud-Edge Collaborative Computing (CECC) framework \cite{he2023priority} combines back-end and edge resources to deliver flexible and efficient services. For example, the EdgeFM system \cite{yang2023edgefm} uses cloud-based Feature Maps to enhance real-time model adaptation, maintaining high accuracy under dynamic conditions. Similarly, the EVOLVE framework \cite{yu2024evolve} advances cloud-edge synergy through multiple cloud-based expert models that refine self-supervised learning on edge devices. Its dynamic expert aggregation mechanism adjusts model weights in real time based on reliability and expertise, effectively managing dynamic data streams and sustaining high performance in complex environments.

\section{IIoT: Evolution Pathway for LMs}\label{sec:Large Models on General Industrial Intelligence}
In IIoT environments, the dynamic nature of data necessitates models that can continuously adapt to evolving scenarios by learning new knowledge. The relevance of continual learning refers to the interconnection between continual learning mechanisms and IIoT environments and applications. Traditional machine learning methods rely on retraining with large datasets, which is both costly and inefficient in IIoT contexts. This inefficiency arises due to the inability of traditional methods to address challenges such as data heterogeneity, resource constraints, and dynamic data streams. Continual learning overcomes these challenges by incrementally integrating new data into existing knowledge, offering a more efficient and cost-effective solution. Furthermore, continual learning enhances model adaptability and scalability. Its applications in IIoT are extensive, spanning areas such as predictive maintenance \cite{10102331,10419797}, intelligent manufacturing, and smart transportation.

\subsection{Theoretical Insights}
\begin{figure}[t]
    \centering
    \includegraphics[width=0.9\linewidth]{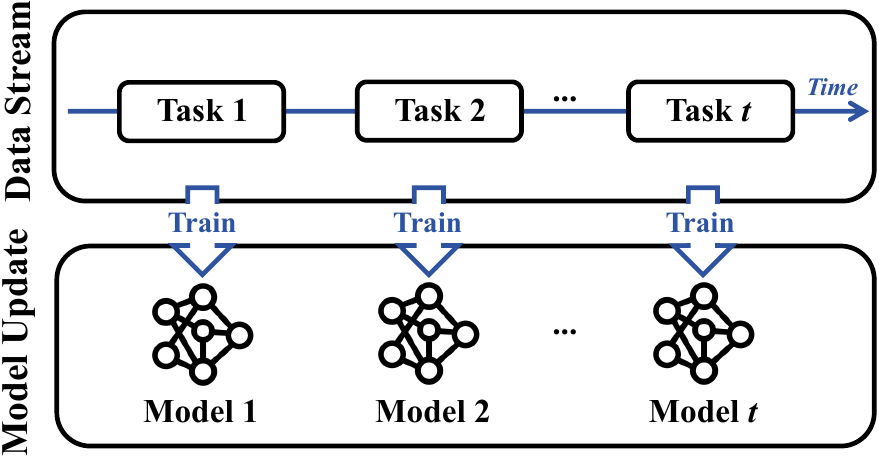}
    \caption{
    General framework of CL. New tasks arrive sequentially, and the model needs to learn them incrementally.
    }
    \label{fig:CL}
\end{figure}

As shown in Fig.~\ref{fig:CL}, CL is a strategy for training machine learning models to adapt to a sequence of changing tasks, each potentially introducing new data distributions \cite{parisi2019continual, hadsell2020embracing}. We define a series of tasks $\{\mathcal{D}_t\}_{t=1}^T$, where each task $\mathcal{D}_t$ consists of a set of data samples and corresponding labels, i.e., $\mathcal{D}_t = \{(\mathbf{x}_i^t, y_i^t)\}_{i=1}^{|\mathcal{D}_t|}$, where $\mathbf{x}_i^t$ is the input data and $y_i^t$ is the corresponding label. The goal in each task stage $t$ is for the model to learn the new task's data while addressing catastrophic forgetting on previous tasks. \looseness=-1

Class-Incremental Learning (CIL) is an important branch of CL. 
In CIL, each task $\mathcal{D}_t$ introduces an independent set of categories, i.e., $y_i^t \in \mathcal{Y}_t$ and $\mathcal{Y}_t \cap \mathcal{Y}_j = \emptyset$ for all $t \neq j$. 
According to the goals of CL, the objective for CIL is to fit a model $f(\cdot)$, and minimize the cumulative expected risk across the entire sequence of tasks:
\begin{equation}
    f^* = \arg \min_{f\in\mathcal{H}} \sum_{t=1}^T \mathbb{E}_{(\mathbf{x}_i^t,y_i^t) \sim \mathcal{D}_t} \left[ \mathbb{I}[f(\mathbf{x}_i^t) \neq y_i^t] \right],
\end{equation}
where $\mathcal{H}$ is the hypothesis space, $\mathbb{I}(\cdot)$ is the indicator function which outputs 1 if the expression holds and 0 otherwise.

Two widely used metrics in continual learning are \textbf{Average Accuracy (ACC)} and \textbf{Forgetting Rate}:

\begin{itemize}
    \item \textbf{ACC:} This metric evaluates the model's overall performance across all tasks after training is completed. It is calculated as the average of the accuracies achieved on all tasks, providing an indication of how well the model retains knowledge of previous tasks while learning new ones.
    
    \item \textbf{Forgetting Rate:} This metric quantifies the degree of performance degradation on earlier tasks as new tasks are learned. It is computed as the difference between the maximum accuracy achieved on a task and the final accuracy on the same task. A lower forgetting rate indicates better retention of past knowledge.
\end{itemize}

\subsection{IIoT for Continual Learning}

The IIoT serves not only as an application domain for continual learning but also as a critical enabling platform. By providing dynamic data streams, edge computing capabilities, and efficient communication networks, IIoT establishes the essential infrastructure for model training, deployment, and real-time updates in continual learning. For example, in the integration of FL and edge computing (MEC), the distributed architecture of IIoT enables cross-device collaboration, supporting knowledge sharing and incremental learning across heterogeneous devices, thereby enhancing model adaptability and efficiency.

Specifically, research in this area focuses on two major directions: the integration of FL with edge computing and the application of continual learning in specific industrial scenarios.

\textbf{Integration of FL with Edge Computing.} IIoT's distributed environment provides a natural foundation for federated learning across devices. Frameworks such as Cross-FCL \cite{9960821} utilize parameter decomposition, global aggregation, and local optimization strategies to retain knowledge and task memory across multiple edge devices, improving model efficiency and accuracy. Additionally, personalized federated learning frameworks \cite{10148063,10628095} are deployed on IIoT devices to enable incremental task learning while reducing communication costs. These methods fully leverage the heterogeneity and distributed nature of IIoT to better adapt models to dynamic environments.

\textbf{Applications in Specific Industrial Scenarios.} IIoT's dynamic data support enables a wide range of applications for continual learning in specific industrial scenarios. For instance, in digital twin synchronization, IIoT provides real-time sensor data streams to maintain synchronization between physical objects and their digital representations \cite{10316647}, using incremental training to handle data drift and real-time updates. In electromagnetic signal classification, methods like SMTC-CL \cite{10730792} address task increments and catastrophic forgetting through selective multi-task coordination. Similarly, in QoE assessment, deep learning-based models leverage continuously incoming data from IIoT to enable efficient evaluation while reducing training complexity \cite{9645165}.

\subsection{Traditional Continuous Learning}
\begin{figure*}[!t]
    \centering
    \begin{tikzpicture}[node distance=0.4cm, minimum width=3cm, font=\scriptsize]

    \definecolor{pink-1}{rgb}{0.976,0.773,0.773}
    \definecolor{pinkpink}{rgb}{1, 0.882,0.902}
    \definecolor{yellow}{rgb}{1, 0.94, 0.86} 
    \definecolor{blue-1}{rgb}{0.86, 0.865, 0.96}
    \definecolor{blue-2}{rgb}{0.56,0.67, 0.9}
    \definecolor{blue-3}{rgb}{0.18,0.33, 0.63}

      \node[draw=blue-3, fill=none, rotate=90, line width = 1pt, rounded corners] (box1) {Continual Learning};
      \node[draw=blue-3,fill=none,right=0.8cm of box1, yshift=-1.5cm, line width = 1pt, rounded corners] (box2) {Replay-based};
      \node[draw=blue-3,fill=none,right=0.8cm of box1, above=1.2cm of box2, line width = 1pt, rounded corners] (box3) {Regularization-based};
      \node[draw=blue-3,fill=none,right=0.8cm of box1, below=1.2cm of box2, line width = 1pt, rounded corners] (box4) {Parameter Isolation};
      \node[draw=blue-3,fill=none,right=of box3, yshift=0.4cm, line width = 1pt, rounded corners] (box5) {Parameter Regularization}; 
      \node[draw=blue-3,fill=none,right=of box3, yshift=-0.4cm, line width = 1pt, rounded corners] (box6) {Functional Regularization};
      \node[draw=blue-3,fill=none,right=of box2, yshift=0.4cm, line width = 1pt, rounded corners] (box7) {Experience Replay};
      \node[draw=blue-3,fill=none,right=of box2, yshift=-0.4cm, line width = 1pt, rounded corners] (box8) {Generative Replay};
    \node[draw=blue-3,fill=none,right=of box4, yshift=0.4cm, line width = 1pt, rounded corners] (box15) {Static Architecture};
      \node[draw=blue-3,fill=none,right=of box4, yshift=-0.4cm, line width = 1pt, rounded corners] (box16) {Dynamic Architecture};

    %Parameter regularization
      \node[draw=blue-3,fill=none, text width=5.5cm, right=of box5, line width = 1pt, rounded corners] (box11) {
      EWC \cite{Kirkpatrick2016OvercomingCF},
      Online EWC \cite{Schwarz2018ProgressC}, 
      SI \cite{Zenke2017ContinualLT}, 
      MAS \cite{Aljundi2017MemoryAS}
    };

    %Functional Regularization
      \node[draw=blue-3,fill=none, text width=5.5cm, right=of box6, line width = 1pt, rounded corners] (box12) {
        LwF\cite{Li2016LearningWF}, 
        DMC \cite{Zhang2019ClassincrementalLV},
        AYT \cite{Szatkowski2023AdaptYT}
      };

    % Experience Replay
      \node[draw=blue-3,fill=none, text width=5.5cm, right=of box7, line width = 1pt, rounded corners] (box13){
      iCaRL \cite{Rebuffi2016iCaRLIC},
    GEM \cite{LopezPaz2017GradientEM},
    A-GEM \cite{Chaudhry2018EfficientLL},
    OML \cite{javed2019meta}
      };

        % Generative Replay
      \node[draw=blue-3,fill=none, text width=5.5cm, right=of box8, line width = 1pt, rounded corners] (box14){
    DGR \cite{Shin2017ContinualLW}, 
    ESGR \cite{He2018ExemplarSupportedGR},
    DDGR \cite{Gao2023DDGRCL},
    DSG \cite{he2024continual},
      };

      % Static Architecture
      \node[draw=blue-3,fill=none, text width=5.5cm, right=of box15, line width = 1pt, rounded corners] (box17){
      PathNet \cite{fernando2017pathnet}, 
      PackNet \cite{mallya2018packnet}, 
      HAT \cite{serra2018overcoming}
      };

      %Dynamic Architecture
      \node[draw=blue-3,fill=none, text width=5.5cm, right=of box16, line width = 1pt, rounded corners] (box18){
      PNNs \cite{rusu2016progressive}, 
      DEN \cite{yoon2017lifelong}
      };
      
      \draw[blue-2] (box1.south) -| ([xshift=-0.25cm]box2.west) -| (box2.west);
      \draw[blue-2] (box1.south) -| ([xshift=-0.25cm]box3.west) -| (box3.west);
      \draw[blue-2] (box1.south) -| ([xshift=-0.25cm]box4.west) -| (box4.west);
      \draw[blue-2] (box3.east) -| ([xshift=-0.25cm]box5.west) -| (box5.west);
      \draw[blue-2] (box3.east) -| ([xshift=-0.25cm]box6.west) -| (box6.west);
      \draw[blue-2] (box2.east) -| ([xshift=-0.25cm]box7.west) -| (box7.west);
      \draw[blue-2] (box2.east) -| ([xshift=-0.25cm]box8.west) -| (box8.west);
      \draw[blue-2] (box4.east) -| ([xshift=-0.25cm]box15.west) -| (box15.west);
      \draw[blue-2] (box4.east) -| ([xshift=-0.25cm]box16.west) -| (box16.west);

      \draw[blue-2] (box5.east) -- (box11.west);
      \draw[blue-2] (box6.east) -- (box12.west);
      \draw[blue-2] (box7.east) -- (box13.west);
      \draw[blue-2] (box8.east) -- (box14.west);
      \draw[blue-2] (box15.east) -- (box17.west);
      \draw[blue-2] (box16.east) -- (box18.west);
      
    \end{tikzpicture}
    \caption{Taxonomy of Traditional CL.}
    \label{fig:three_type_cl}
\end{figure*}

As shown in Fig.~\ref{fig:three_type_cl}, existing continual learning surveys generally classify methods into the following categories:

\begin{itemize}
    \item \textbf{Regularization-based methods}: These mitigate catastrophic forgetting of previous tasks by adding regularization terms to the loss function.
    \item \textbf{Replay-based methods}: These alleviate catastrophic forgetting within a limited memory buffer by using data from previous tasks.
    \item \textbf{Architecture-based methods}: These allocate independent parameters for new task data, for example, through parameter allocation, model partitioning, or modular network models.
\end{itemize}

\subsubsection{Regularization-Based Methods}\label{sec:Regularization-based Methods}

Regularization-based methods aim to mitigate catastrophic forgetting by incorporating regularization terms into the training process. These methods can be broadly categorized into two subtypes: parameter regularization and functional regularization.

\textbf{Parameter Regularization Methods.}
\begin{figure}[t]
    \centering
    \includegraphics[width=0.75\linewidth]{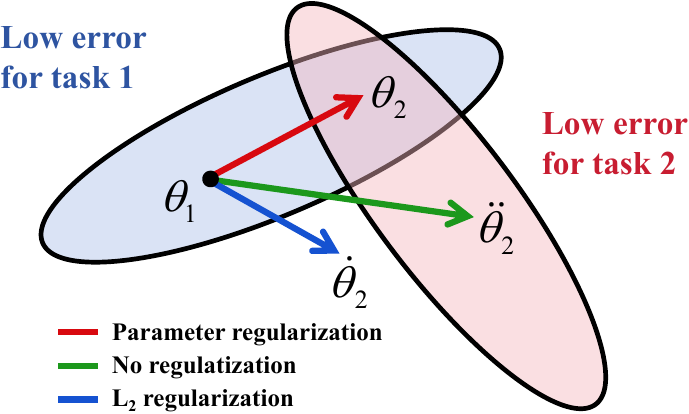}
    \caption{Illustration of the effect of parameter regularization across two tasks.}
    \label{fig:EWC}
\end{figure}
Parameter regularization constrains changes to critical parameters, balancing the plasticity-stability trade-off. The regularization term penalizes significant deviations in important parameters during new task training, as illustrated in Fig.~\ref{fig:EWC}. For a task $t$, given the model parameters $\theta$, the regularized loss $\tilde{\mathcal{L}}_{t}(\theta)$ is expressed as \cite{Zenke2017ContinualLT}:
\begin{equation}
    \tilde{\mathcal{L}}_{t}(\theta) = \mathcal{L}_{t}(\theta) + c \sum_{i} \Omega_{i}^{t} \left(\theta_{i} - \theta_{t-1, i}^{*}\right)^{2},
\end{equation}
where $\mathcal{L}_{t}(\theta)$ represents the task-specific loss for task $t$, $\Omega_{i}^{t}$ is the importance matrix of the $i$-th parameter for task $t$, $\theta_{t-1, i}^{*}$ is the $i$-th parameter value after task $t-1$, and $c$ is a hyperparameter controlling the balance between stability and plasticity.

Elastic Weight Consolidation (EWC) \cite{Kirkpatrick2016OvercomingCF} was the first method to use parameter regularization, estimating parameter importance using the Fisher Information Matrix (FIM). From a Bayesian perspective, the conditional probability $p(\theta \mid \mathcal{D}_{1:t})$ for parameters after task $t$ can be expressed as:
\begin{equation}
    p(\theta \mid \mathcal{D}_{1:t}) \propto p(\theta) \prod_{i=1}^{t} p(\mathcal{D}_{i} \mid \theta) \propto p(\theta \mid \mathcal{D}_{1:t-1}) p(\mathcal{D}_{t} \mid \theta),
\end{equation}
where $p(\theta \mid \mathcal{D}_{1:t-1})$ encapsulates information from previous tasks \cite{Kirkpatrick2016OvercomingCF}. However, this posterior distribution is often intractable and requires approximations such as Laplace approximation, variational inference, or moment matching. Using Laplace approximation, the posterior is approximated as a multivariate Gaussian:
\begin{equation}
    p(\theta \mid \mathcal{D}_{1:t-1}) \approx \mathcal{N}(\theta; \theta_{t-1}^{*}, \Lambda_{t-1}^{-1}),
\end{equation}
where $\Lambda_{t-1}$ is the precision matrix. EWC approximates $\Lambda_{t-1}$ using a diagonal FIM, resulting in the following regularized loss:
\begin{equation}
    \tilde{\mathcal{L}}_{t}(\theta) = \mathcal{L}_{t}(\theta) + c \sum_{i} F_{i} \left(\theta_{i} - \theta_{t-1, i}^{*}\right)^{2},
\end{equation}
where $F_{i}$ is the FIM of the $i$-th parameter.

Subsequent efforts have refined parameter regularization. SI \cite{Zenke2017ContinualLT} measures parameter importance based on its contribution to the loss decay along the training trajectory, while MAS \cite{Aljundi2017MemoryAS} calculates importance via the sensitivity of output predictions to parameter changes using unlabeled data.

\textbf{Functional Regularization Methods.}
Functional regularization primarily leverages knowledge distillation to retain knowledge from prior tasks. The central idea is to preserve the network's output predictions for previous tasks while training on new tasks \cite{DeLange2019ACL}. Learning without Forgetting (LwF) \cite{Li2016LearningWF} serves as a foundational method in this category. It achieves this by employing two loss components: one to guide task-specific learning for the new task and another to align the current predictions with those generated by the original network on prior tasks. This ensures that the model retains its ability to perform previous tasks without catastrophic forgetting.

Instead of relying on detailed equations, LwF utilizes a distillation-based approach, where the network's softened probabilities (calculated using a temperature scaling factor) are used to balance the performance on both old and new tasks. The relative importance of these components is controlled through weighting factors, which ensure that past knowledge is preserved without hindering the learning of new tasks.

Recent advancements in functional regularization have extended its capabilities. For instance, DMC \cite{Zhang2019ClassincrementalLV} incorporates double distillation techniques to mitigate biases between new and old classes, thereby improving class-incremental learning performance. Similarly, AYT \cite{Szatkowski2023AdaptYT} enables synchronous updates of teacher and student models, making it particularly effective in exemplar-free class-incremental learning scenarios. These methods reflect the continuous evolution of functional regularization strategies to better address the challenges of continual learning.

\subsubsection{Replay-Based Methods}\label{sec:Replay-based Methods}

Replay-based methods mitigate catastrophic forgetting by replaying data from previous tasks during the learning of new tasks. Depending on how data is replayed, these methods are categorized into two main types: experience replay and generative replay.

\textbf{Experience Replay Methods.}  
Experience replay methods store a subset of data samples from previous tasks in a memory buffer $\textit{B}$ and combine them with data from the current task during training. This approach effectively preserves knowledge from earlier tasks and prevents catastrophic forgetting.

Based on Learning without Forgetting (LwF), iCaRL \cite{Rebuffi2016iCaRLIC} introduces an exemplar-based memory mechanism for class-incremental learning (CIL). It maintains exemplar sets $\mathcal{P}$, which are updated whenever new classes are learned. Each exemplar set $P$ for a specific class is defined as $P = \left ( p_{1}, p_{2}, ..., p_{m}\right )$, where $m$ denotes the capacity of the set. The total memory capacity $K$ is distributed across all observed classes $t$, such that $m = K/t$.

The selection of exemplars is performed iteratively. For the $i$-th exemplar in the set, $p_{i}$ is chosen as:
\begin{equation}
    p_{i} \leftarrow \underset{x \in X}{\operatorname{argmin}}\left\|\mu-\frac{1}{i}\left[\varphi(x)+\sum_{j=1}^{i-1} \varphi\left(p_{j}\right)\right]\right\|,
\end{equation}
where $X=\{ x_1, x_2, ..., x_n \}$ represents the data samples for a class, $\varphi(\cdot)$ is the feature function, and $\mu$ is the class mean, computed as:
\begin{equation}
    \mu = \frac{1}{n} \sum_{x \in X} \varphi(x).
\end{equation}
At the start of training for class $t$, the memory buffer $\mathcal{P} = \{ P_{1}, P_{2}, ..., P_{t-1} \}$ stores exemplars from previously observed classes.

Beyond sampling strategies, other methods focus on effectively leveraging stored data. Gradient Episodic Memory (GEM) \cite{LopezPaz2017GradientEM} imposes inequality constraints on the losses of previous tasks to avoid forgetting, addressing violations through gradient projection. OML \cite{javed2019meta}, a meta-learning approach, trains a deep Representation Learning Network (RLN) to learn sparse and robust features, enhancing its ability to retain knowledge across tasks.

\textbf{Generative Replay Methods.}
Generative replay methods employ a generative model to synthesize pseudo-data that represents previous tasks, thereby reducing the dependency on large memory buffers typically required in experience replay \cite{Shin2017ContinualLW, he2024continual}. As illustrated in Fig.~\ref{fig:generator}, there are three common types of generative models used in generative replay: Variational Autoencoders (VAEs), Generative Adversarial Networks (GANs), and Diffusion Models.

GANs have been extensively applied in generative replay due to their ability to produce high-quality and realistic data \cite{He2018ExemplarSupportedGR,Shin2017ContinualLW}. Recently, diffusion models \cite{Gao2023DDGRCL, he2024continual} have also been utilized for data generation, leveraging their robustness and capability to capture complex data distributions. These methods effectively mitigate memory constraints and ensure knowledge retention, making them particularly suitable for real-world continual learning applications.

\subsubsection{Parameter Isolation Methods}\label{sec:Parameter Isolation Methods}

Parameter isolation methods address catastrophic forgetting by allocating specific parameters within a network for each task. These methods can be categorized into static architecture methods and dynamic architecture methods, depending on whether the network structure remains fixed during training.

\textbf{Static Architecture Methods.}  
Static architecture methods operate under a fixed network structure, allocating specific subsets of parameters for task-specific learning.

PathNet \cite{fernando2017pathnet} introduces pathways that consist of active modules in each layer of the network. Each task is assigned its optimal pathway, which becomes fixed after training. This prevents further modifications to the parameters associated with that task, effectively preserving the acquired knowledge.

PackNet \cite{mallya2018packnet} leverages pruning techniques to allocate parameters for new tasks. Initially, the network learns the parameters for the first task, after which a subset of the learned weights is pruned (set to zero). The pruned weights are then reallocated and trained for subsequent tasks. For each task $t$, after training, pruning is applied to the weights allocated for task $t$, and these pruned weights are retrained for task $t+1$. To preserve knowledge from previous tasks, weights associated with earlier tasks remain frozen. Masks are maintained to indicate the active parameters relevant to each task, enabling the network to selectively activate specific parameters during task inference.

HAT \cite{serra2018overcoming} introduces a hard attention mechanism that employs task-specific masks at the unit level. These masks constrain parameter updates during new task learning, ensuring the retention of knowledge from prior tasks while enabling task-specific learning.

\textbf{Dynamic Architecture Methods.}  
Dynamic architecture methods allow the network to expand as new tasks are introduced, retaining the parameters associated with previously learned tasks.

Progressive Neural Networks (PNNs) \cite{rusu2016progressive} address catastrophic forgetting by creating a new network for each task. Knowledge transfer is facilitated through lateral connections, which enable the new network to utilize features learned by earlier networks. This approach ensures the preservation of previously acquired knowledge while supporting task-specific learning for new tasks.

Dynamic Expandable Networks (DEN) \cite{yoon2017lifelong} utilize group-sparsity regularization to determine the necessity of adding new neurons at each layer for a given task. By expanding the network only when required, DEN minimizes unnecessary network retraining and maintains efficiency.

\subsection{Emerging Continual with LPMs}
\begin{figure*}[t]
    \centering
    \includegraphics[width=0.9\linewidth]{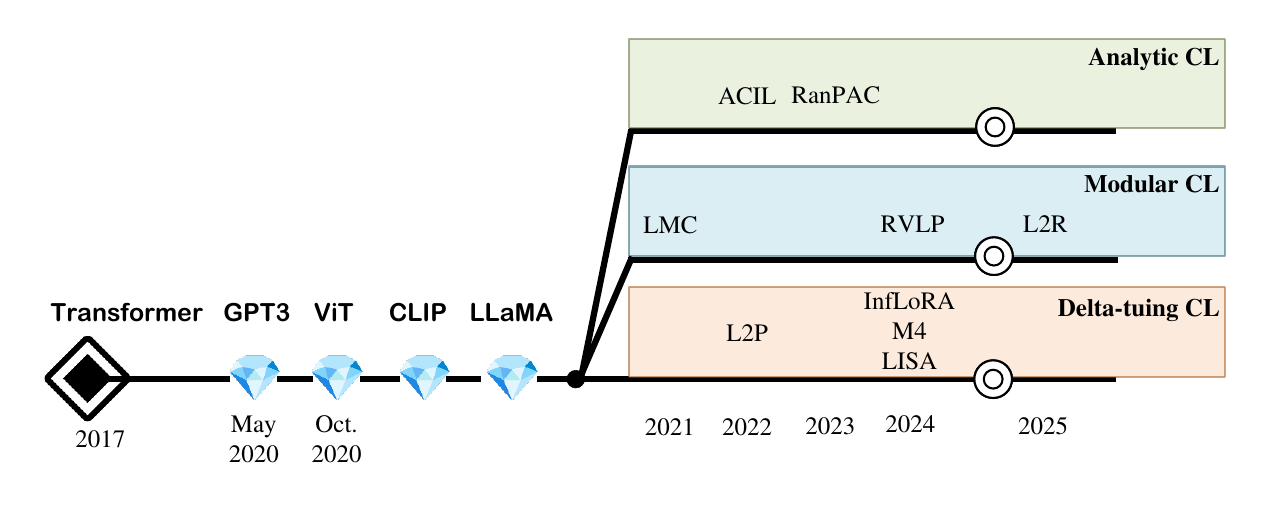}
    \caption{
    Development timeline of large pre-trained models based continual learning.
    }
    \label{fig:outline}
\end{figure*}

As shown in Fig.~\ref{fig:outline}, the field of continual learning with large pre-trained models (LPMs) has evolved significantly in recent years, driven by innovative approaches designed to address the challenges of adapting to new tasks while preserving prior knowledge. This section introduces three prominent paradigms—Delta-tuning CL, Modular CL, and Analytic CL—and highlights their unique mechanisms and contributions.

\subsubsection{Delta-tuning CL}

Delta-tuning continual learning (Delta-tuning CL) is an emerging paradigm that emphasizes the principles of \textbf{minimal modification and high efficiency}. This method freezes most parameters of LPMs and introduces a small number of trainable parameters to adapt to new tasks \cite{wang2022learning,liang2024inflora,pan2024lisa,yuan2024mobile}. 
This design contrasts sharply with traditional full-parameter fine-tuning methods. Delta-tuning CL preserves the generalization ability of LPMs to the greatest extent while significantly reducing computational and memory costs. This approach is particularly suitable for resource-constrained scenarios and environments with complex and dynamic tasks \cite{jia2022visual}.

\subsubsection{Modular CL}

Modular continual learning (Modular CL) is a novel framework that uses modular structures and routing mechanisms to represent and compose knowledge dynamically \cite{ostapenko2021continual,araujo2024learning,qu2025introducing}. This framework divides knowledge into independent modules. Each module serves as a distinct functional or structural unit. The routing mechanism selects or combines modules dynamically based on the characteristics of the input. This design creates an adaptive system tailored to the requirements of specific tasks.

\subsubsection{Analytic CL}
\begin{figure}[t]
    \centering
    \includegraphics[width=1\linewidth]{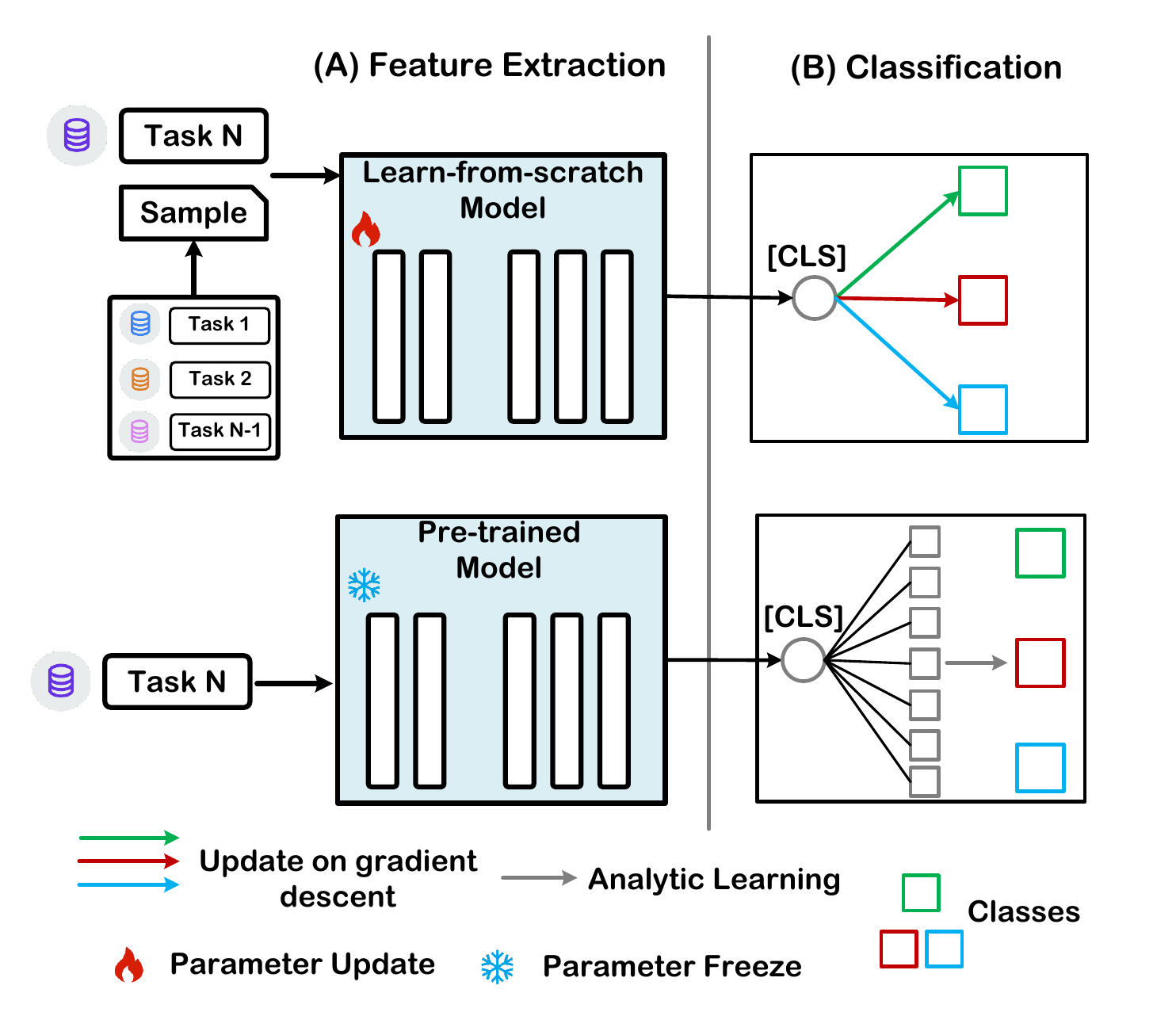}
    \caption{
    Analytic CL Based on Pre-trained Models vs. Traditional Replay-based CL.
    }
    \label{fig:analytic learning}
\end{figure}

As shown in Fig.~\ref{fig:analytic learning}, Analytic continual learning (Analytic CL) is a continual learning paradigm centered on analytic methods, utilizing explicit computation of model parameter updates to achieve efficient retention of historical knowledge. Unlike traditional optimization methods based on gradient descent, Analytic CL employs linear algebra and analytical solutions to update model parameters.

The initial representative method, Analytic Class-Incremental Learning (ACIL) \cite{zhuang2022acil, zhuang2023gkeal} focuses on analytic learning by recursively updating weight matrices and regularized feature autocorrelation matrices (RFAuM) \cite{zhuang2021blockwise}. ACIL achieves theoretical absolute memorization by reconstructing results identical to joint learning using only data from the current task without requiring historical data.

As analytic methods advance, Random Projections and Class-Prototype Strategy (RanPAC) \cite{mcdonnell2024ranpac} integrates LPMs and feature space optimization, expanding the practical applications of Analytic CL. RanPAC employs frozen random projection layers and nonlinear random projections to map features from LPMs into higher-dimensional spaces, improving feature linear separability. Additionally, the method uses class-prototype strategies and second-order feature statistics, such as covariance matrices, to represent class distributions.

\section{Outlook and Future GII}\label{sec:Outlook and Future Opportunitie}

The IIoT is not merely a data pipeline but serves as an intelligent infrastructure for the development of LMs. With its vast data resources, low-latency high-bandwidth networks, and edge computing capabilities, the IIoT provides essential support for the training, deployment, and continuous evolution of LMs. As illustrated in Fig.~\ref{fig:iiot}, the lifecycle of LMs within the IIoT consists of four key stages. The following sections outline future research directions and technical advancements anticipated in each of these stages.

\subsection{Data Generation in IIoT}\label{sec:future Data Generation}

Data generation, particularly diffusion models \cite{ho2020denoising} and other generative AI techniques, are playing an increasingly critical role in IIoT. These methods can produce high-quality synthetic data to address data scarcity, enhance model robustness, and ensure data privacy. 

\textbf{Lesson Learnt}: Diffusion models demonstrate great potential in small-sample scenarios, but their high computational cost requires further optimization to suit resource-constrained edge devices \cite{10398474}. For time-series data \cite{10380228} and image/video data \cite{yang2024drivearena}, generation techniques must account for temporal and multimodal characteristics while incorporating domain-specific prior knowledge for better optimization. Moreover, data generation can simulate diverse fault scenarios, providing richer datasets for model training and evaluation.

\subsection{Model Training in IIoT}\label{sec:Learning with Restricted Computational Resources}

Federated learning and continual learning are set to become the dominant approaches for model training in IIoT. Federated learning effectively addresses challenges such as data silos and privacy concerns, while continual learning enables models to adapt to dynamically changing environments. Recent research in CL increasingly focuses on applying models in resource-constrained scenarios. This trend is expected to further highlight computational efficiency challenges and propose solutions in future developments \cite{prabhu2023computationally}.

\textbf{Lesson Learnt:}  
The application of federated learning in IIoT faces challenges related to communication efficiency, model consistency, and security. Asynchronous federated learning, model compression, and blockchain technologies have been shown to effectively mitigate these issues. For continual learning, addressing catastrophic forgetting and knowledge transfer remains critical. While traditional replay- and regularization-based methods have limitations, modular and analytic approaches exhibit greater potential.

Edge-based LMs can achieve fast, near real-time analysis and response. However, due to the resource constraints of edge nodes and end devices, they struggle to support such large models \cite{10508191, 9933792}. Therefore, integrating end, edge, and cloud computing to leverage their respective advantages is essential to compensate for resource limitations at any single level.

\subsection{Model Routing and Merging in IIoT}
As the scale and complexity of LLMs continue to grow, model routing and merging technologies are becoming increasingly important. These techniques enable the deployment of different model components across various devices and dynamically combine models based on task requirements, achieving more efficient and flexible model deployment and execution. For instance, the ``AI Flow" framework dynamically evaluates task demands and network conditions to allocate inference tasks optimally, improving resource utilization \cite{shao2024ai}. On the other hand, model merging integrates models trained on different devices, combining diverse data and knowledge to significantly enhance model performance and generalization. For example, the TAKFL \cite{morafahtowards} framework employs independent distillation, self-regularization, and task-level arithmetic knowledge integration to achieve efficient and secure model merging.

\textbf{Lesson Learnt:}  
Dynamic task allocation and model integration can improve system efficiency and robustness; however, they require careful consideration of device heterogeneity and resource constraints to ensure optimal performance. Furthermore, context-aware routing and performance-based routing are two commonly used strategies that can enhance task assignment by dynamically adapting to environmental conditions and resource availability. Additionally, model merging must address challenges such as model conflicts and knowledge integration to effectively combine knowledge from diverse models while maintaining system consistency and accuracy.

\subsection{Continuous Evolution in IIoT}
The continual evolution of LLMs in the IIoT represents a critical area of research. Models must continuously acquire new knowledge and skills to adapt to ever-changing environments and task requirements. In terms of model reuse, IIoT provides distributed computational capabilities, enabling techniques such as duplication \cite{chen2021bert2bert}, stacking \cite{gong2019efficient}, and weight combination \cite{wang2023learning} to be efficiently implemented in resource-constrained environments. Additionally, IIoT supports knowledge distillation to initialize large Transformer models, significantly reducing the need for data samples and stabilizing optimization processes. Even under extreme data scarcity, model reuse can effectively reproduce scaling laws \cite{bahri2021explaining}, ensuring efficient and reliable learning.

\textbf{Lesson Learnt:}  
Existing continual learning methods primarily focus on addressing catastrophic forgetting in classification tasks but need to be extended to other task types, such as regression, prediction, and control. Furthermore, new continual learning approaches must be developed to enhance learning efficiency, generalization, and interpretability.  
Zero-waste continual learning emphasizes maximizing the utilization of existing resources and computational capacity, making it a pivotal direction for the future of continual learning. For instance, through the use of a knowledge pool \cite{wang2022learning}, models can rapidly access task-relevant knowledge to adapt to new tasks. By employing fine-tuning techniques such as joint reparameterization, pre-trained models can efficiently transfer and integrate this knowledge, optimizing learning efficiency and adaptability.

\section{Conclusion}
\label{sec:conclusion}
This paper explores the evolution of the Industrial Internet of Things (IIoT) from a traditional data pipeline to an intelligent infrastructure that supports the full lifecycle of industrial large models. First, it introduces various types of industrial large models and their typical applications, focusing on fault diagnosis, predictive maintenance, planning, and control.
Next, the paper examines the rich data resources provided by IIoT, highlighting how techniques such as data freshness, data denoising, selection, and generation ensure data quality to support model training and fine-tuning.
It then analyzes the infrastructural role of IIoT in model training, emphasizing how low-latency and high-bandwidth networks enable efficient model learning and real-time adaptation.
Subsequently, the paper explores the advantages of deploying large models in IIoT environments, particularly how modular design and routing mechanisms enhance system flexibility and adaptability to meet diverse industrial needs.
Finally, the paper examines the continual learning mechanisms of large models, focusing on how traditional and pre-trained model-based continual learning methods ensure long-term effectiveness and adaptability in dynamic industrial environments.
Through this comprehensive discussion, the survey provides a thorough perspective and in-depth analysis of the application of large models in IIoT and their contribution to advancing General Industrial Intelligence (GII).

In the future, we will focus on providing a more detailed exploration of the application scenarios of industrial large models in key stages such as research and development, production and manufacturing, testing and validation, and operational management, highlighting their potential value and practical applications.
\balance

\section{Acknowledgements}
We would like to extend our special thanks to Ruyi Huang, Longya Xiao, Yihao Wang, Qiurui Zhang, Chenguang Ma, Xiang Zhang, Di Fang and Run He from the Shien-Ming Wu School of Intelligent Engineering, South China University of Technology, China, for their invaluable support in providing case studies, data, and materials related to industrial applications, which greatly enriched the depth and practical relevance of this paper.

\bibliographystyle{IEEEtran}  
\bibliography{IEEEabrv,references}
\end{CJK}
\end{document}